\documentclass[12pt]{article}
\usepackage[utf8]{inputenc}
\usepackage[ruled, vlined]{algorithm2e}
\usepackage{amsmath}
\usepackage{amssymb}
\usepackage{biblatex}
\usepackage{booktabs}
\usepackage{caption}
\usepackage{cancel}
\usepackage{color}
\usepackage[export]{adjustbox}
\usepackage{graphicx}
\usepackage{hyperref}
\hypersetup{ colorlinks=true,
    linkcolor=red,
    filecolor=cyan,      
    urlcolor=blue,}
\usepackage{float}
\usepackage[margin=1in]{geometry}
\usepackage{nomencl}
\usepackage{titlesec}
\graphicspath{ {Images/} }
\renewcommand{\baselinestretch}{1.2}
\newcommand\tab[1][1cm]{\hspace*{#1}}
\usepackage{setspace}
\usepackage{listings}
\numberwithin{equation}{section}
\usepackage{xcolor}

\begin{document}
\begin{titlepage}
\newcommand{\HRule}{\rule{\linewidth}{0.5mm}}
\center
\textsc{\LARGE Technion - Israel Institute of Technology}\\[0.8cm]
{\Large \textbf{Faculty of Aerospace Engineering}}\\[2.3cm]
\begin{figure}[H]
\centering
\includegraphics[scale=0.2]{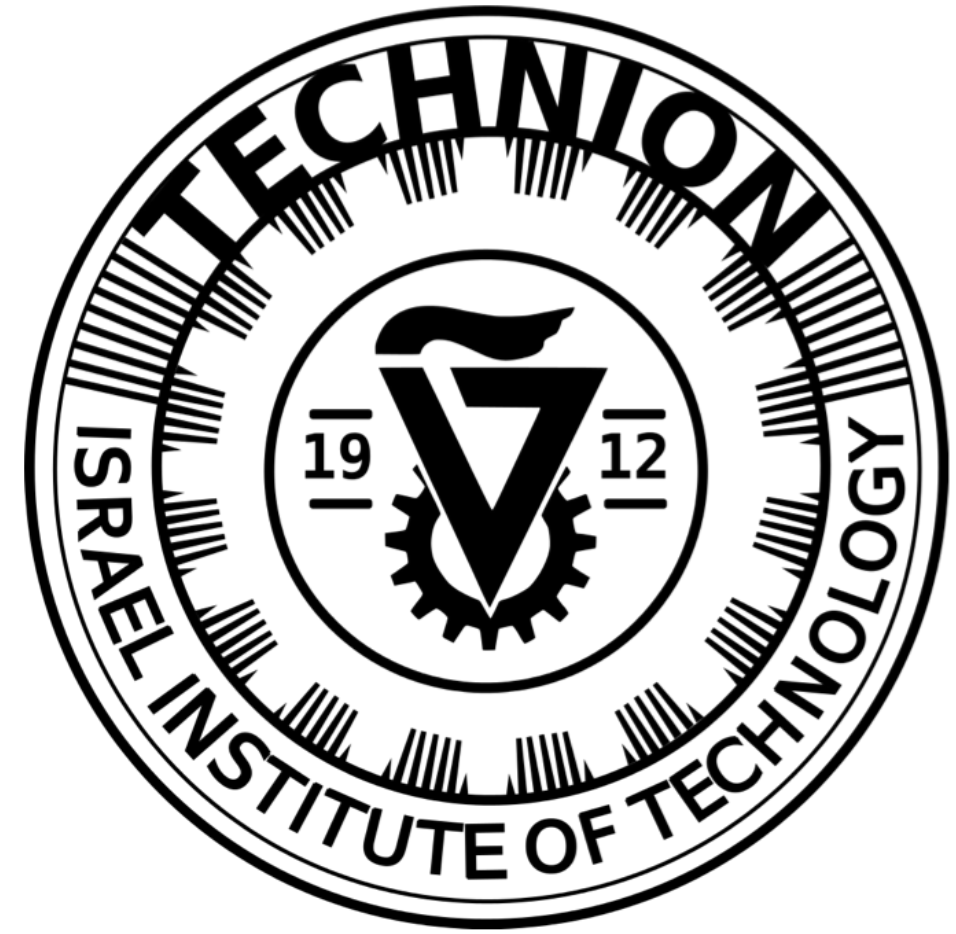}
\end{figure}

\vspace*{50px}
\textbf{{\Large A Study of a Genetic Algorithm for Polydisperse \\[0.5cm] Spray Flames}}

\vspace*{80px}

\HRule \\[0.3cm]
\textbf{Final Project Report Towards M.Eng in Aerospace Engineering}\\[0.2cm]
\textbf{Advisor : Prof. Barry Greenberg}\\[0.2cm]
\textbf{Submitted by : Daniel Engelsman}\\[0.2cm]
\HRule \\
\end{titlepage}

\begin{flushleft} 
\pagenumbering{gobble} 
\begingroup
\hypersetup{linkcolor=black}
\renewcommand{\baselinestretch}{0.85}\normalsize
\tableofcontents 
\renewcommand{\baselinestretch}{1.0}\normalsize
\endgroup 				
\newpage

\pagenumbering{arabic} 

\subsection*{Abstract}  \addcontentsline{toc}{subsection}{Abstract}
Modern technological advancements constantly push forward the human-machine interaction. Nowadays, finding an application whose algorithm does not utilize a Machine Learning (ML) methods, is quite rare. The reason for that is their capability of solving abstract problems that were so far not even expressible, and thus remained enigmatic. 

Evolutionary Algorithms (EA) are an ML subclass inspired by the process of natural selection - "\textit{Survival of the Fittest}", as stated by the Darwinian Theory of Evolution. The most notable algorithm in that class is the Genetic Algorithm (GA) - a powerful heuristic tool which enables the generation of a high-quality solutions to optimization problems. In recent decades the algorithm underwent remarkable improvement, which adapted it into a wide range of engineering problems, by heuristically searching for the optimal solution.

Despite being well-defined, many engineering problems may suffer from heavy analytical entanglement when approaching the derivation process, as required in classic optimization methods. Therefore, the main motivation here, is to work around that obstacle.

In this piece of work, I would like to harness the GA capabilities to examine optimality with respect to a unique combustion problem, in a way that was never performed before. To be more precise, I would like to utilize it to answer the question : "What form of an initial droplet size distribution (iDSD) will guarantee an optimal flame ?"

To answer this question, I will first provide a general introduction to the GA method, then develop the combustion model, and eventually merge both into an optimization problem.

\subsection*{Acknowledgments} \addcontentsline{toc}{subsection}{Acknowledgments}
First and foremost, I would like to thank Ms. Debbie Warril. \\ Without you I wouldn't have made it this far. As simple as that. Thank you.

I would also like to pay my gratitude to the project supervisor - Professor Barry Greenberg. For the professional academic guidance and kind hearted support all along the way.

Last but not least, I would like to thank my beloved parents and family, just for being who you are. I believe we are lucky people in this world.

\newpage


\section*{Nomenclature} \addcontentsline{toc}{subsection}{Nomenclature}

\subsubsection*{Latin symbols}
\begin{tabular}{ p{1.5cm} p{10cm} p{3cm} }
$B$ & Dimensionless reaction & $[-]$ \\
$B_{i, i+1}$ & $i$-th integral coefficient & $[1/\text{sec}]$ \\
$C_{i}$ & $i$-th integral coefficient & $[1/\text{sec}]$ \\
$C_P$ & Specific heat at constant pressure & $[J/(Kg \cdot K)]$ \\
$c$ & Half inner channel size (normalized) & $[-]$ \\
$d$ & Droplet diameter & $[m]$ \\
$D_g$ & Mass diffusion coefficient & $[m^2/sec]$ \\
$E$ & Evaporation rate & $[m^2/sec]$ \\
$F$ & Drag acceleration & $[m/sec^2]$ \\
$K$ & Thermal diffusion coefficient & $[m^2/sec]$ \\
$\bar{E}$ & Normalized evaporation rate & $[m^2]$ \\
$L$ & Half inner channel size & $[m]$ \\
$m$ & Mass fraction & $[-]$ \\
$\dot{m}$ & Mass flux & $[Kg / (m^2 \cdot sec)]$ \\
$N$ & Number of sections & $[-]$ \\
$n$ & Droplet size probability function & [Droplets] \\
$q$ & Heat flux & $[J/(m^2 \cdot sec)]$ \\
$R$ & Half external channel size & $[m]$ \\
$\tilde{R}$ & Droplet volumetric change rate & $[m^3/sec]$ \\
$S$ & Mass source / sink element & $[1/sec]$ \\
$\bar{S}$ & Heat source / sink element & $[K/sec]$ \\
$t$ & Time & $[sec]$ \\
$\tilde{T}$ & Temperature & $[K]$ \\
$T$ & Dimensionless temperature & $[-]$ \\
$T_{max}$ & Normalized maximum tip flame temperature & $[-]$ \\
$U_g$ & Characteristic flow velocity & $[m/sec]$ \\
$V$ & Oxidizer's initial mass fraction & $[-]$ \\
$v$ & Droplet volume & $[m^3]$ \\
$x$ & Cartesian coordinate perpendicular to the flow & $[m]$ \\
$y$ & Cartesian coordinate parallel to the flow & $[m]$ \\
\end{tabular}

\subsubsection*{Greek symbols}
\begin{tabular}{ p{1.5cm} p{10cm} p{3cm} }
$\alpha$ & Coefficient of the integral property & $[-]$ \\
$\gamma$ & Normalized mass fraction & $[-]$ \\
$\gamma_d$ & Normalized mass fraction (droplets) & $[-]$ \\
$\gamma_F$ & Normalized mass fraction (fuel) & $[-]$ \\
$\gamma_O$ & Normalized mass fraction (oxidizer) & $[-]$ \\
$\gamma_T$ & Normalized mass fraction (temperature @ S-Z) & $[-]$ \\
$\delta$ & Normalized initial mass fraction (fuel) & $[-]$ \\
$\Delta$ & Damköhler number for evaporation & $[-]$ \\
$\Delta_i$ & Integral coefficient of $i$-th section & $[-]$ \\
$\eta$ & Normalized coordinate parallel to the flow & $[-]$ \\
$\eta_{max}$ & Normalized maximum flame height & $[-]$ \\
$\Lambda$ & Normalized latent heat & $[-]$ \\
$\xi$ & Normalized coordinate perpendicular to the flow & $[-]$ \\
$\rho$ & Density & $[Kg / m^3]$ \\
$\nu$ & Stoichiometric coefficient & $[-]$ \\
\end{tabular}

\subsubsection*{Shortcuts}
\begin{tabular}{ p{1.5cm} p{10cm} p{3cm} }
Da & Damköhler number for reaction & $[-]$ \\
DoF & Degree of Freedom &  \\ 
EA & Evolutionary Algorithms &  \\
F/O & Fuel / Oxidizer ratio & $[-]$ \\
GA & Genetic Algorithm &  \\
iDSD & initial Droplet Size Distribution &  \\
Le & Lewis number & $[-]$ \\
LHS & Left Hand Side &  \\
ML & Machine Learning &  \\
Pe & Péclet number & $[-]$ \\
rev. & Reversal (point) &  \\
RHS & Right Hand Side &  \\
SMD & Sauter Mean Diameter & $[m]$ \\
S-Z & Schwab-Zeldovich transformation & \\
We & Webber number & $[-]$ \\
\end{tabular}
\newpage

\section{Introduction} 

\subsection{Background} \label{Background}

The GA consists of 3 mechanisms that reflect the natural selection process (survival of the fittest), where the fittest is selected for producing the next generation's offspring [\ref{c6}].

\textbf{Natural selection} - The individual's / parent probability to be selected to produce the offspring of the next generation. Its genetic variation determines its survival chances, and thus points which chromosomes are to be preserved and multiplied between generations. In nature, it's caused by forced competitive interaction between different populations and individuals, that rewards the fittest among them (best intelligence, physique etc.). 

\textbf{Crossover} - The recombination of the genetic information of two parents (in nature known as mating). Better individuals will participate in the production of the next generation, such that the last generation will hold the best former genetic qualities. Each \underline{single} crossover (out of hundreds) is subjected to a \underline{random} partitioning, across every generation.

\begin{figure}[H]
\centering
\includegraphics[scale=0.29]{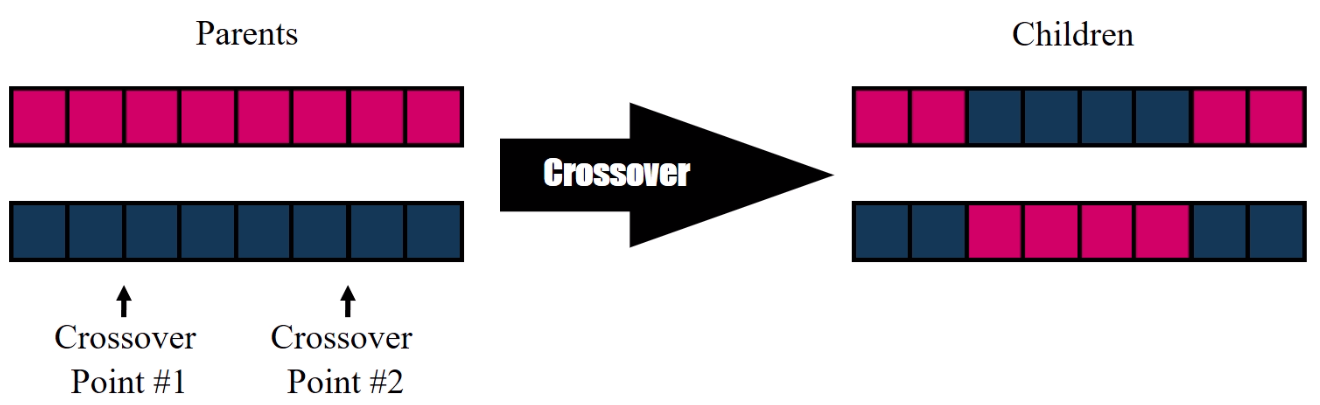}
\end{figure}
\textbf{Mutation} - Random genetic alteration during the recombination process. The bigger the sample space is, the wider the diversity becomes, and so do the chances for new improved features. A positive / negative feature caused by a mutation will be reflected in the individual's fitness quality, resulting in higher / lower chances to transfer its genes. \newline
\begin{figure}[H]
\centering
\includegraphics[scale=0.375]{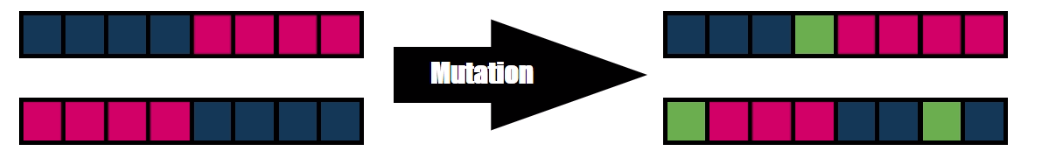}
\end{figure}

$\bullet$ \ \textbf{Chromosome} - the individual's set of properties that represent a candidate solution. As such, the chromosome is subjected to modifications at every generation.

$\bullet$ \ \textbf{Population} - set of $n$-random candidate solutions, given an optimization problem.

The above mechanisms are applied in a loop, where each iteration (=generation) the fittest individuals are extracted and go through genetic recombination. This process continues until a termination criterion is met, and the best candidate solution is received. From left to right is the evolution process from 1st random initialization until the 78th generation : \vspace{5mm}

\begin{figure}[H]
\centering
\includegraphics[width=1.13\linewidth, center]{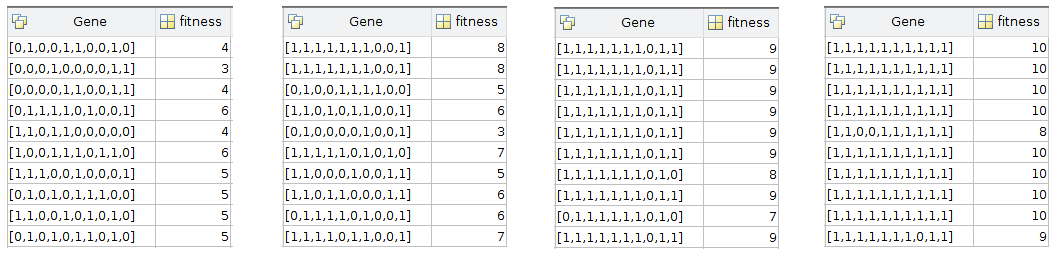}
\end{figure}
Consider the above fitness function to be an $\ell_1$-norm \hspace{0.75mm} : \,$f_{GA}(x_{i}^t) = \| x_i^t \|_1 = \sum_j | x_{i,j} |$. \\ The optimal candidate solution at generation $t$ means  : \,$\text{arg} \, \max_{x^t} \, f_{GA}(X^t)$ . \\
Note the fitness (score) evolution from $t=0$ until $t=78$ across all of the population.

As seen, the chromosomes are consisted of atomic sequences named Genes. Assuming an arbitrary chromosome that's composed of $n$ genes we get \,$x_i \in \mathbb{R}^n$\,, \,whereas the different chromosomes can be seen as a points in the \,$\mathbb{R}^{n}$\, space, whose genes are coordinates. 

The GA then, is responsible for finding the best performance among all candidates, as they are measured by a fitness function, or equivalently as they are projected onto a metric axis. The optimal solution (minimum / maximum) is actually the candidate whose score performs best in the $\mathbb{R}^{n+1}$, namely the closest to the global extremum (see \hyperlink{app_extremum}{Appendix A}).

Classic optimization techniques utilize a closed form objective (=fitness) functions that are conveniently differentiable. By calculating the roots of their first and second derivatives, one can extract solutions in the form of minima, maxima or a saddle point [\ref{c_optima}].

However, in complex analytical cases (as in ours), one would rather work around that tiring derivation process which can impose significant challenges, and implement instead solution oriented heuristic methods. For illustration, consider the following 3D function : 
\begin{align*}
f(x, y, z) = \frac{sin(x-x_0)}{x-x_0} \cdot \frac{sin(x-y_0)}{x-y_0} \cdot \Big( -z^3 + z_0 \Big) \quad ; \quad ( x_0, y_0, z_0 ) = ( 10, 10, 1 )
\end{align*}

\begin{figure}[H]
\centering
\includegraphics[width=0.9\linewidth, center]{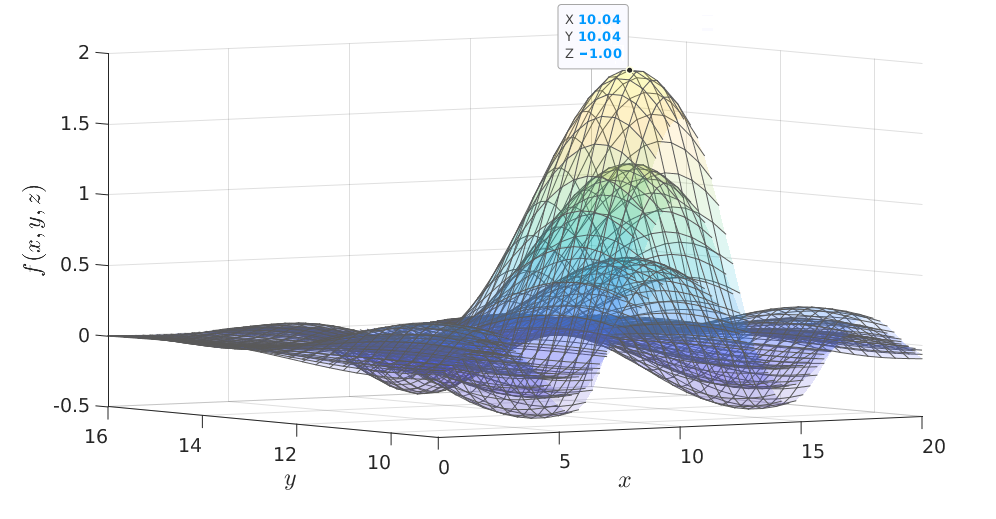}
\caption*{Domain : $\{ 0 \leq x \leq 20 \ , \ 0 \leq y \leq 20 \ , \ -1 \leq z \leq 1 \}$}
\end{figure}
The GA manages to find a global solution, being slightly dependent on the mesh resolution.

The individual's score at a given generation is measured by its performance to a desired fitness function. That function acts as a comparison measure, where the best candidate is the one whose score is optimal. That optimality can be either minimum or maximum, depending on the problem's nature - concave, convex or non-convex (see \hyperlink{app_convex}{Appendix B}).

One of the main advantages of the GA is its indifference to the internal workings of the fitness function, namely it refers to it as a "black box", evaluating different points \,$f_{GA}(x)$\, due to its coding methodology.

\subsection{GA in the service of combustion problems}

As introduced above, and detailed thoroughly \href{https://www.cs.colostate.edu/~genitor/MiscPubs/tutorial.pdf}{here}$^{[\ref{c7}]}$, the GA are powerful heuristic search methods, successfully used to find optimal or near-optimal solutions in many complex design spaces. Early implementations of the GA in context of combustion were made back in the 90's, as Runhe Huang (1995) [\ref{ga0}] showed an implementation on a combustion control problem. Instead of learning a control action for every point encountered, a GA was used to learn control actions for a set of limited number of prototype states, and afterwards applying nearest neighbour matching to extract the optimal rule. 

Danielson et al. (1998) [\ref{ga1}], presented the GABSys (GA Bond Graph System), an optimization tool that utilized a 2-stroke combustion engine model : 
\begin{figure}[H]
\centering
\includegraphics[width=0.915\linewidth, center]{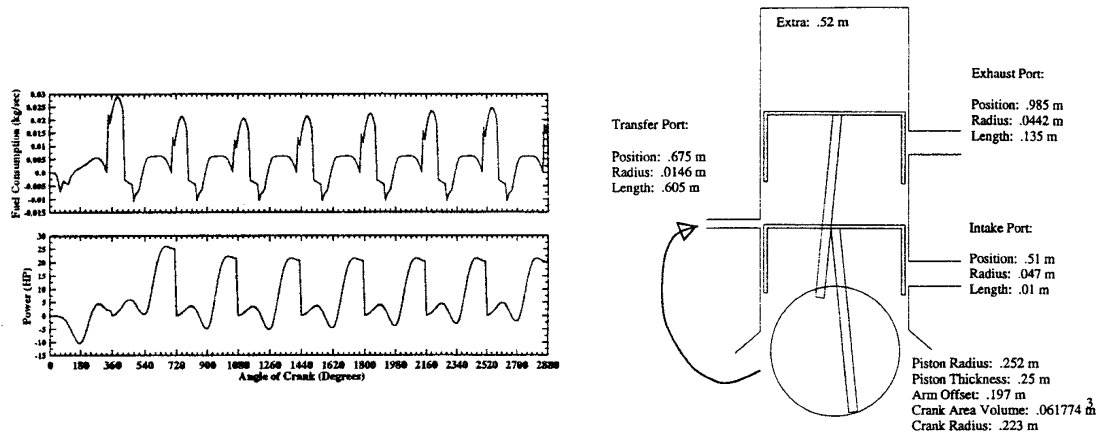}
\end{figure}
They parametrized several related factors (geometric, thermodynamic etc.), and expressed the objective (fitness) function in terms of fuel consumption, power etc. (along cycles), such that eventually a performance space could be spanned and satisfy local extrema : "Although there is no guarantee that the result is optimal, the resulting engine is still very impressive considering that after fewer than 3,000,000 designs from a domain of 1.329$\cdot 10^{36}$, the GA selects a constructable fuel-efficient engine design".

Polifke et al. (1998) [\ref{ga2}] used the GA for combustion reactive mechanism to carry out the subtle optimization process, with a minimum human effort. Harris et al. (1999) [\ref{f11}] used the GA for determining the optimal reaction rate parameters of the O/F mixture.

Vossoughi $\&$ Rezazadeh (2005) [\ref{f12}] introduced a multi-objective GA for an engine control unit, where the objective functions were tailored to the calibration parameters in sought of optimal configuration. Quite similarly, Rose et al. (2009) [\ref{ga3}] implemented the GA on a gas-exchange system of combustion engine, producing a significantly higher power output than was achieved through a basic manual optimization procedure. 

Shtauber \& Greenberg (2010) [\ref{ga33}] conducted a wide study of polydisperse
spray diffusion flames. By analytical and numerical investigation, they have shown the iDSD influence upon the flame properties and its sensitivity to extinction. It is worth mentioning that this project is considerably a continued work on Shtauber's thesis, but focuses primarily on finding the optimal flame properties, using the GA.

Sikalo et al. (2015) [\ref{f1}] described an automatic method for the optimization of reaction rate constants of reduced reaction mechanisms. Based on GA, the technique aimed at finding new reaction rate coefficients that minimize the error introduced by the preceding reduction process. The error was defined by an objective function that covers regions of interest where the reduced mechanism may deviate from the original mechanism.

Kaplan et al. (2015) [\ref{f2}] presented a general approach for developing an automated procedure to determine optimal reaction parameters for a simplified model to simulate flame acceleration and deflagration-to-detonation (DDT) in a methane-air mixture. The laminar flame profile was computed using reaction parameters in a 1D Navier-Stokes code, and matched the profile obtained by a detailed chemical reaction mechanism. 

Pan et al. (2018) [\ref{ga4}] utilized the GA in a boiler combustion control system, to optimize the bias coefficients that maintain the excess air ratio at the optimal combustion interval under variable load conditions (the blue plot) :
\begin{figure}[H]
\centering
\includegraphics[width=1.04\linewidth, center]{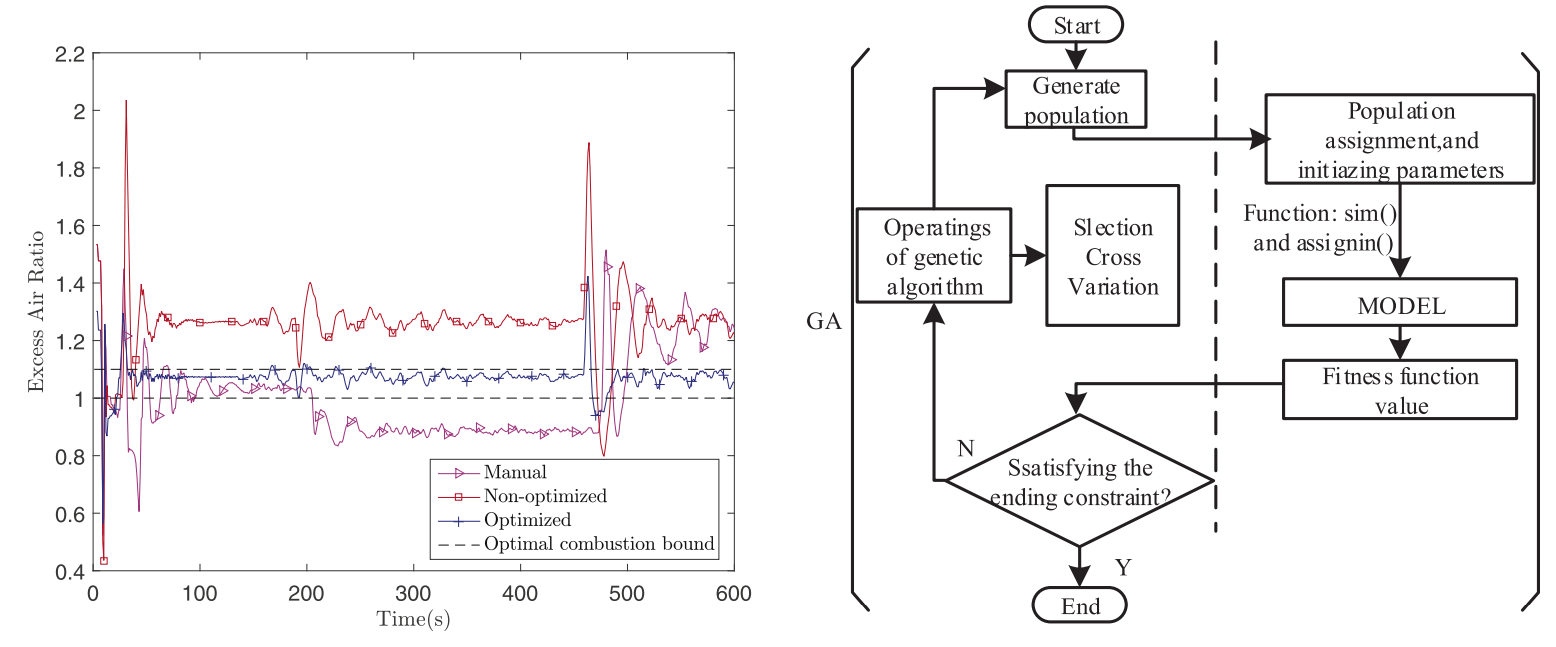}
\end{figure}

Liu et al. (2019) [\ref{ga5}] presented a GA implementation on dual diesel/natural injection parameters, where the indicated specific fuel consumption (NO$x$ and CH$_4$) emissions are selected as the optimization objectives. Similarly, Zhao et al. (2019) [\ref{ga7}] focused on reducing unburned carbon by optimizing operating parameters via a novel high-efficient GA, which was experimentally validated.

\subsubsection*{Summary}

Many of the above researchers had no relation to the combustion physical aspects. Instead, they used it as a convenient optimization framework, for its convenient modelability and being experimentally validable. The main implementations were :

( $ \circ $ ) \ Holistic analysis of an engineering systems (mainly combustion configuration) e.g internal combustion or spark ignition engines, heat exchangers, chemical reactors etc.

( $ \circ $ ) \ State space representation of control systems and attempt to optimize a desired variable (power, emission, efficiency, fuel consumption etc.)

Applying the GA on big frameworks may provide high-level understanding of the engine efficiency, emission aspects or different dynamic profiles. However, smaller focus areas that actually comprise the problem's inner core, might be lost. More precisely, they are not even expressed in the cost function, and are thus overridden by macroscopic interests.

\subsection{Intention statement}

In the absence of any research that applied the GA with a well-defined combustion model, I aim to focus on smaller scopes of interest, primarily on the flame characteristics. 

At first, by being able to express the temperature field and investigating its reactions to a wide range of parameters. To that end, factors like the maximum flame height and the maximum tip flame temperature will be serve as indicators.

Afterwards, I would like to gain control on the GA model by being able to execute optimization schemes in a growing complexity order, by either extreme chemical scenarios or by maximizing the degree of freedom (DoF).

Finally, when full integration is achieved, the GA will be harnessed for the sake of optimization scenarios in order to shed light on the principal factors that may optimize the current combustion model. These steps comprise the current piece of work, in a way that was not conducted before, especially on the seam between GA and combustion. 
I aim to innovate by investigating the cause and effects evoked as a result of the optimality, and validate them according to the literature.

\newpage

\section{The Combustion Model}
In this section I will develop the governing equations describing the mathematical model of the polydisperse spray diffusion flame. That is by presenting the underpinning assumptions, equations, normalization, boundary conditions and full solution. Afterwards I will validate them with a set of results, which will be followed by discussions.  

\subsubsection*{The big picture}

Based on the classic flame model of Burke \& Schumann (1928) [\ref{ga8}], the F/O interaction is separated by a steady state diffusion flame, in a laminar parallel co-flow :

\begin{figure}[H]
\centering
\includegraphics[scale=0.425]{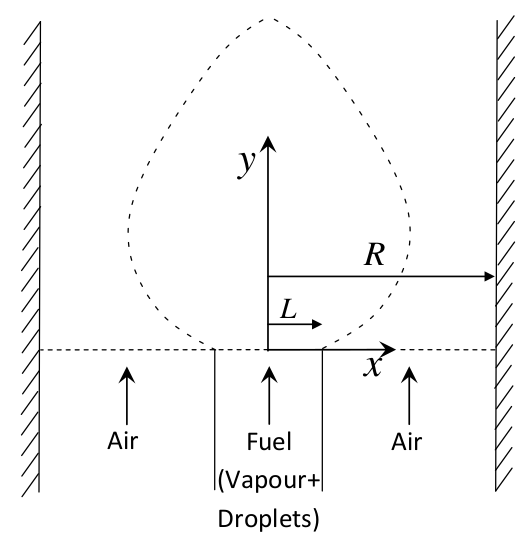}
\end{figure}

This model was further elaborated by Greenberg (1989) [\ref{ga9}] by assuming that the liquid fuel droplets were homogeneously suspended in an inert gas stream.

Using Tambour (1985) [\ref{ga10}] sectional approach for describing the spray polydispersity, in addition to the droplets evaporation rate, it is necessary to refer to the droplets different sizes, as a result of the iDSD and the evaporation rate mechanism ( \,$d^2 \, law$ ).

\newpage

\subsection{The Spray Equation}
Tambour’s sectional approach assumes discrete distribution of the droplets, whose most fundamental size of is called Monomer. Its size will dictate the field’s resolution, and it expresses the spray as a probabilistic function of the droplets :
\begin{align} \label{2_sections}
n = n(t, x, y, v) \hspace{4mm} \underset{continuous}{\rightarrow} \hspace{4mm} n(t, x, y, v) \cdot dx \cdot dy \cdot dv
\end{align}
Each droplet of the spray is indicated by\ $j \in \mathbb{N}$\ that denotes the number of monomers carried inside (e.g number of molecules in a droplet) :
\begin{align} 
\frac{d n_j}{dt} = -E_j n_j + E_{j+1} n_{j+1} \quad , \quad j =1, 2, ...
\end{align}
$E_j$ indicates the evaporation rate of the single $j$-th droplet, such that it is not dependent on the environment's temperature, but on the droplet’s size, based on $d^2\, law$. We’ll divide the spray into N sections and define an integral property (IP) for the $i$-th section :
\begin{align} \label{2_Q_i}
Q_i(t, x, y, v) \triangleq \int_{v_{L_i}}^{v_{H_i}} \alpha v^{\hat{\gamma}} n(t, x, y, v) dv
\end{align}
The volume defines the $i$-th section and is bounded within $v \in [v_{L_i} , v_{H_i} )$ . The $\hat{\gamma}$ coefficient defines the property whereas $\hat{\gamma} = \{0 , 1 , \frac{2}{3} \}$ based on the IP’s dependence on the number of droplets, their volume or their surface area. Then $\alpha$ is set into a desirable property (e.g density) such that the integration would yield the $i$-th mass section. Further he showed :
\begin{align}  \label{2_dQ_i}
\frac{d Q_i}{dt} = -C_i Q_i + B_{i, i+1} Q_{i+1} \quad , \quad i=1, \ldots, N
\end{align}
Where the general integral coefficients are :
\begin{align} 
B_{i, i+1} = \Big( \frac{v_{H_i}}{v_{L_{i+1}}} \Big)_{\textbf{I}}^{\hat{\gamma}} \, \frac{E(v_{H_i})}{v_{H_{i+1}} - v_{L_{i+1}} } \quad , \quad B_{N, N+1}=0  \label{2_B_i} \\
C_{i} = \Big( \frac{v_{H_{i-1}}}{v_{L_{i}}} \Big)^{\hat{\gamma}}_{\textbf{II}} \, \frac{E(v_{L_i})}{v_{H_{i}} - v_{L_{i}} } + \frac{1}{v_{H_{i}} - v_{L_{i}} } \int_{v_{L_i}}^{v_{H_i}} \frac{1}{v^{\hat{\gamma}}} E(v)\ \underset{\textbf{III}}{dv^{\hat{\gamma}}} \label{2_C_i}
\end{align}

I. \hspace{2.45mm} Expresses $Q_i$’s growth \hspace{5.75mm} by droplets \underline{addition} from $i + 1 \rightarrow i$ section. \\
II. \hspace{1.0mm} Expresses $Q_i$’s diminution by droplets \underline{downgrade} from the $i$-th section. \\
III.\, Expresses $Q_i$’s diminution by droplets \underline{evaporation} in the $i$-th section.

Using the general conservation equation given by Williams (1985) [\ref{ga11}] :
\begin{align} 
\frac{ \partial n}{ \partial t} + \frac{ \partial}{ \partial v} ( \underbrace{\tilde{R}}_{ \frac{\partial v}{\partial t}} n ) + \nabla \cdot ( U_d n ) + \nabla_{U_d} \cdot ( \underbrace{F}_{  \underset{ \text{accel.}}{\text{Drag}}} n ) = \Gamma
\end{align} 
Where $\Gamma$ is a source of droplets resulted from the collision rate. However, we should recall some of the prior assumptions made in the thesis [\ref{ga33}] :
\begin{align}
\underbrace{\nabla_{U_{d}} \cdot(F n)}_{v_{d r o p}=v_{g a s}}=0 \tab \underbrace{\frac{\partial n}{\partial t}}_{ \underset{\text {state}}{\text{steady}} }=0 \tab \underbrace{\Gamma}_{ \underset{\text {collision}}{\text{negligible}} }=0 \tab \underbrace{U_{d}=U_{0_{g}}}_{\underset{\text {collision}}{\text{negligible}} }=0
\end{align}  
Using the sectional approach based on these assumptions, we get the following equation :
\begin{align}
\nabla \cdot\left(U_{0_{g}} Q_{i}\right)=-C_{i} Q_{i}+B_{i, i+1} Q_{i+1} \quad, \quad i=1, \dots, N
\end{align}
Plugging $\left(\hat{\gamma}=1, \, \alpha=\rho_{d} / \rho_{T o t}\right)$ in (\ref{2_Q_i}) we get the mass fraction equation $\left(Q_{i}=m_{d_{i}}\right)$
\begin{align}
U_{0_{g}} \frac{\partial m_{d_{i}}}{\partial y}=-C_{i} m_{d_{i}}+B_{i, i+1} m_{d_{i+1}} \quad, \quad i=1, \ldots, N
\end{align}
The \underline{continuous} sectioning $\left(\begin{array}{c}d_{H_{i-1}} \rightarrow d_{L_{i}} \\ d_{H_{i}} \rightarrow d_{L_{i+1}}\end{array}\right)$ of the integral coefficients (\ref{2_B_i}, \ref{2_C_i}) becomes :
\begin{align}
B_{i, i+1}= \frac{3}{2} E\left[\frac{d_{L_{i+1}}}{d_{H_{i+1}}^{3}-d_{L_{i+1}}^{3}}\right]  \quad , \quad i=1, \ldots, N \\
C_{i} =\frac{3}{2} E\left[\frac{3 d_{H_{i}}-2 d_{L_{i}}}{d_{H_{i}}^{3}-d_{L_{i}}^{3}}\right] \quad, \quad i=1, \ldots, N 
\end{align}

\newpage

\subsubsection{Boundary conditions}

The droplets are described in a 1st order equation whose BC are the liquid fuel mass
fraction at the channel's exit. Away from the nozzle ("far field") the polydispersity is assumed to be homogeneous:
\begin{align} 
m_{d_{i}}=m_{\text{Tot, fuel}} \cdot \begin{cases}
\delta_{i} \ , & 0 \leq x \leq L \\
0 \ , & L<x \leq R
\end{cases} \quad , \quad i=1, \ldots, N
\end{align}
Applying the following normalizations :
\begin{align}
(\xi, \eta, c) \ \triangleq \ \left( \frac{x}{R} \ , \frac{y D_{g}}{U_{0_{g}} R^{2}} \ , \frac{L}{R} \right) \label{2_axes_norm} \\
\left(\gamma_{d_{i}}\right) \ \triangleq \ \left(m_{d_{i}} / m_{\text {Tot, fuel}}\right) \\
\left(\psi_{i}, \Delta_{i}\right)\ = \ \frac{R^{2}}{D_{g}}\left(B_{i}, C_{i}\right)
\end{align}
And we get the dimensionless spray equation :
\begin{align} 
\frac{\partial \gamma_{d_{i}}}{\partial \eta}=-\Delta_{i} \gamma_{d_{i}}+\psi_{i} \gamma_{d_{i+1}} \label{2_d_gamma}
\end{align}
Whereas the integral coefficients are defined as :
\begin{align}  
\bar{E}=E\left(\frac{R^{2}}{D_{g}}\right) \quad \underset{1 \leq i \leq N}{\Delta_i}=\frac{3 E}{2}\left(\frac{3 d_{H_{i}}-2 d_{L_{i}}}{d_{H_{i}}^{3}-d_{L_{i}}^{3}}\right) \quad \underset{i \leq i \leq N-1}{\psi_i} =\frac{3 \bar{E}}{2}\left(\frac{d_{i+1}}{d_{H_{i+1}}^{3}-d_{L_{i+1}}^{3}}\right)  \label{2_coeffs}
\end{align}
And the dimensionless BC are :
\begin{align} 
\gamma_{d_{i}}=\frac{m_{d_{i}}}{m_{Tot, fuel}}= \begin{cases}
\delta_{i}, & 0 \leq \xi \leq c \\
0, & c<\xi \leq 1  \end{cases} \quad , \quad  i=1, \ldots, N  \label{2_gamma}
\end{align}

\newpage

\subsubsection{Analytical solution}

The spray equation (\ref{2_d_gamma}) contains the coefficients (\ref{2_coeffs}) and the compatible BC (\ref{2_gamma}). Since subsequent equations are mutually dependent, we'll propose an iterative approach :
\begin{align} 
\gamma_{d_{j}}=\sum_{i=j}^{N} \Omega_{i j} e^{-\Delta_{i} \eta} \quad ; \quad \frac{\partial \gamma_{d_{j}}}{\partial \eta}=-\sum_{i=j}^{N} \Delta_{i} \Omega_{i j} e^{-\Delta_{i} \eta} \quad ; \quad  \underset{(i>j)}{\Omega_{i j}}=\frac{\psi_{j}}{\Delta_{j}-\Delta_{i}} \Omega_{i, j+1} \label{2_dimen_less}
\end{align}
$\Omega_{i j}$ is an influence coefficient. By plugging inside the dimensionless spray equation (2.17) :
\begin{align} 
-\sum_{i=j}^{N} \Delta_{i} \Omega_{i j} e^{-\Delta_{i} \eta}=-\Delta_{j} \sum_{i=j}^{N} \Omega_{i j} e^{-\Delta_{i}\eta}+\psi_{j} \sum_{i=j+1}^{N} \Omega_{i, j+1} e^{-\Delta_{i} \eta}
\end{align}
Applying solution on the BC :
\begin{align} 
\gamma_{d_{j}}(0)=\sum_{i=j}^{N} \Omega_{i j}= \delta_{j} \\
\Omega_{j j}=\sum_{i=j}^{N} \delta_{i}-\sum_{i=j+1}^{N} \Omega_{i j}=\gamma_{d_{j}}(0)-\sum_{i=j+1}^{N} \Omega_{i j}\ , \ (j<N) \\
\Omega_{N N}=\sum_{i=j}^{N} \delta_{i}=\gamma_{d_{N}}(0)
\end{align}
The above solution is approximated as continuous, despite the mass fraction being discontinuous as droplets may join or leave the $i$-th section (=discrete phenomenon). 

This approximation allows the analytical solution as it presumes that the average spray injection may contain up to hundred thousands of droplets. That way, joining of a \underline{single} droplet from larger section, is negligible, as it is smaller by several order of magnitudes.

To sum up the droplets solution :
\begin{align} 
\gamma_{d_{j}}=\sum_{i=j}^{N} \Omega_{i j} e^{-\Delta_{i} \eta} \\
\underset{(i>j)}{\Omega_{i j}} = \frac{\psi_{j}}{\Delta_{j}-\Delta_{i}} \Omega_{i, j+1} \\
\Omega_{j j}=\gamma_{d_{j}}(0)-\sum_{i=j+1}^{N} \Omega_{i j}=\delta_{j}-\sum_{i=j+1}^{N} \Omega_{i j}
\end{align}

\newpage

\subsection{The Gaseous Phase Equation}

According to Fick's 1st law for diffusion, the mass flux is linear with its spatial gradient :
\begin{align} 
\dot{m}_{A}=-D_{A} \nabla \rho_{A}
\end{align}
Combining it with the continuity equation provides its 2nd law, AKA the diffusion equation and refers to the concentration change as a function of time. Using that, the gaseous phase
equation in terms of mass fraction will be expressed as :
\begin{align} 
U_{0_{A}} \frac{\partial m_{A}}{\partial y}=D_{A}\left(\frac{\partial^{2} m_{A}}{\partial x^{2}}+\frac{\partial^{2} m_{A}}{\partial y^{2}}\right)+S_{A}
\end{align}
This equation is valid for both gaseous fuel and oxidizer where $S_{A}$ is a sink / source of element $A$ in terms of rate. Assuming equal diffusion coefficients and equal velocities for both gaseous fuel and oxidizer :
\begin{align} 
\text { Fuel : } \ U_{0_{g}} \frac{\partial m_{g, \text { fuel }}}{\partial y}=D_{g} \Big( \frac{\partial^{2} m_{g, \text { fuel }}}{\partial x^{2}}+\frac{\partial^{2} m_{g, \text { \text{fuel} }}}{\partial y^{2}} \Big)+S_{g, \text { fuel-reac }}+S_{d, \text { fuel }} \label{3_fuel} \\
\text { Oxidizer : } \ U_{0_{g}} \frac{\partial m_{O_{2}}}{\partial y}=D_{g} \Big( \frac{\partial^{2} m_{O_{2}}}{\partial x^{2}}+\frac{\partial^{2} m_{O_{2}}}{\partial y^{2}} \Big)+S_{O_{2}-\text{reac}} \hspace{29mm} \label{3_oxidizer}
\end{align}
Using Schwab-Zeldovich (S-Z) transformation will help us uniting both equations :
\begin{align} 
m \triangleq m_{\text{g, fuel}}-m_{O_{2}} / \nu
\end{align}
Where the stoichiometric ratio $(\nu)$ is originated at : \\ \vspace{3mm}
\hspace{45mm} Fuel \ + \ $\underline{\nu}$ Oxygen \ $\rightarrow$ \ Heat \ + \ Products 

And the reactant elements are active only at the reaction zone such that :
\begin{align} 
S_{g, \text { fuel-reac }}=S_{O_{2}, \text{ fuel-reac }} / \nu
\end{align}
So by implementing the transform : (\ref{3_fuel})\,-\,(\ref{3_oxidizer}) / $v$  we can get rid of the reactants and the nonlinear reaction rate term :
\begin{align}
U_{0_g} \frac{\partial m}{\partial y}=D_{g} \Big( \frac{\partial^{2} m}{\partial x^{2}}+\frac{\partial^{2} m}{\partial y^{2}} \Big)+S_{d}
\end{align}
$S_{d}$ is the source element expressing the droplets evaporation rate and their contribution to the gaseous phase. Using this we can find the flame shape by solving for $\underline{m=0}$ .
\newpage


\subsubsection{Boundary conditions}

The gaseous phase equation requires 2 BC for each axis. The {BC} in the channels' exit contain diffusive flux elements resulting from the mass fraction gradient :
\begin{align} 
y=0 \ : \ m_{g, \text{fuel}}-\frac{D_{g}}{U_{0_{g,  \text{fuel}}}} \frac{\partial m_{g,{\text{fuel}}}}{\partial y}= \begin{cases}
m_{\text{Tot, fuel}}\left(1-\sum_{i=1}^{N} \delta_{i}\right) & , \quad 0 \leq x \leq L \\
0 & , \quad L \leq x \leq R
\end{cases}
\end{align}
Similarly, we get in the oxidizer equation :
\begin{align} 
y=0 \ :  \hspace{7mm} m_{O_{2}}-\frac{D_{g}}{U_{0_{9,\text{fuel}}} } \frac{\partial m_{O_{2}}}{\partial y}=\begin{cases}
0 & , \quad 0 \leq x \leq L \\
m_{O_{2}}(y=0) & , \quad L \leq x \leq R
\end{cases}  \hspace{21mm}
\end{align}
And by using S-Z transformation :
\begin{align} 
m-\frac{D_{g}}{U_{0_{g,\text{fuel}}}} \frac{\partial m}{\partial y}= \begin{cases}
m_{ \text{Tot, fuel}}\left(1-\sum_{i=1}^{N} \delta_{i}\right) & , \quad 0 \leq x \leq L \\
-m_{O_{2}}(y=0) / \nu & , \quad L \leq x \leq R 
\end{cases}
\end{align} 
Applying the following assumptions :
\begin{align} 
\underbrace{\frac{\partial m}{\partial x}}_{\text {Symmetry }}\bigg|_{x=0} ^{y \geq 0}=0 \quad \underbrace{\frac{\partial m}{\partial x}}_{  \underset{\text {channel}}{\text{Impenetrable}}} \bigg|_{x=R} ^{y \geq 0}= 0 \quad \underbrace{\frac{\partial m}{\partial y}}_{  \underset{\text {equilibrium}}{\text{Thermodynmic}}} \bigg|_{0 \leq x \leq R} ^{y \rightarrow \infty}=0
\end{align} 
Axes normalization is similar as before $(2.14),$ but we'll add :
\begin{align} 
(\gamma, V)=\frac{\left(m, m_{O_{2}}(y=0) / \nu\right)}{m_{\text{Tot, fuel}}}
\end{align} 
And we get the dimensionless gaseous phase equation :
\begin{align} 
\frac{\partial \gamma}{\partial \eta}=\frac{\partial^{2} \gamma}{\partial \xi^{2}}+\frac{1}{P e^{2}} \cdot \frac{\partial^{2} \gamma}{\partial \eta^{2}}+\bar{S}_{d}
\end{align} 
Where $\left(P e, \bar{S}_{d}\right)$ stands for Peclet number and the normalized source element :
\begin{align} 
P e=\frac{U_{0_{d}} R}{D_{g}} \tab \bar{S}_{d}=\frac{S_{d} R^{2}}{D_{g}\, m_{ \text{Tot, fuel}}}
\end{align} 

\newpage 
The source element in the gaseous fuel equation equals to the droplets evaporation rate and stems from the overall rates over $\mathrm{N}$ sections :
\begin{align} 
\bar{S}_{d} \triangleq \sum_{j=1}^{N} \Delta_{i} \gamma_{d_{i}}-\psi_{i} \gamma_{d_{i+1}}
\end{align} 
Such that the dimensionless gaseous phase equation :
\begin{align} 
\gamma-\frac{1}{P e^{2}} \frac{\partial \gamma}{\partial \eta}= 
\begin{cases}
1-\sum_{i=1}^{N} \delta_{i} & , \quad 0 \leq \xi \leq c \\
-V & , \quad c \leq \xi \leq 1
\end{cases} \label{3_gas_phase}
\end{align} 
And the dimensionless BC are :
\begin{align} 
\frac{\partial \gamma}{\partial \xi} \bigg|_{\xi=0,1}^{\eta \geq 0}= 0 \tab \frac{\partial \gamma}{\partial \eta} \bigg|_{0 \leq \xi \leq 1}^{\eta \rightarrow \infty}=0
\end{align} 

\subsubsection{Analytical solution}
Using S-Z transform, the normalized mass fraction fulfils $: \quad \gamma \triangleq \gamma_{F}-\gamma_{O}$ .

We can solve it as a sum of the following equations :
\begin{align} 
\text { Homogenous } &: \ \frac{\partial \gamma_{h}}{\partial \eta}=\frac{\partial^{2} \gamma_{h}}{\partial \xi^{2}}+\frac{1}{P e^{2}} \cdot \frac{\partial^{2} \gamma_{h}}{\partial \eta^{2}} \\
\text { Particular } &: \ \frac{\partial \gamma_{p}}{\partial \eta}=\frac{\partial^{2} \gamma_{p}}{\partial \xi^{2}}+\frac{1}{P e^{2}} \cdot \frac{\partial^{2} \gamma_{p}}{\partial \eta^{2}}+\bar{S}_{d}
\end{align} 

\subsubsection*{Homogeneous solution}
Both equations will be solved using the separation of variables method - $\gamma_{h}=f_{h}(\eta) \cdot g_{h}(\xi)$
\begin{align} 
\frac{d f_{h}}{d \eta} \cdot g_{h}=f_{h} \cdot \frac{d^{2} g_{h}}{d \xi^{2}}+\frac{1}{P e^{2}} \cdot \frac{d^{2} f_{h}}{d \eta^{2}} \cdot g_{h} \\
\left(\frac{d f_{h}}{d \eta}-\frac{1}{P e^{2}} \cdot \frac{d^{2} f_{h}}{d \eta^{2}}\right) / f_{h}=\frac{d^{2} g_{h}}{d \xi^{2}} / g_{h}=\textit{ Const. } \triangleq-\alpha
\end{align} 

Solve $g_{h}(\xi)$
\begin{align} 
\frac{d^{2} g_{h}}{d \xi^{2}}=-\alpha g_{h} \quad \rightarrow \quad g_{h}=C_{1} \sin (\sqrt{\alpha} \xi)+C_{2} \cos (\sqrt{\alpha} \xi)
\end{align} 
Solve $f_{h}(\xi)$ :
\begin{align} 
\frac{d^{2} f_{h}}{d \eta^{2}}-(P e^{2}) \frac{d f_{h}}{d \eta}-(\alpha P e^{2}) f_{h}=0
\end{align} 

Guess the exponential solution of the form of $f_{h}=C_{3} e^{C_{4} \eta}$ and plug it above :
\begin{align} 
\text { Plugging } \Rightarrow \quad \cdot  (\frac{1}{C_{3} e^{C_{4} \eta}} ) \quad \Rightarrow \quad C_{4}^{2}- (Pe^{2} ) C_{4}- (\alpha Pe^{2} )=0 \\
C_{4_{1,2}}= \frac{1}{2} [ (Pe^{2} ) \pm \sqrt{(Pe)^{4}+4 \alpha(Pe)^{2}}] \underbrace{<}_{\text {fulfill BC}} 0 \\
C_{4}=\frac{ (Pe^{2} )}{2} \bigg[1-\sqrt{1+\frac{4 \alpha}{(Pe)^{2}}} \bigg]  \triangleq \mathbf{q}_{n}
\end{align} 
And finally we get the homogeneous equation solution :
\begin{align} 
\gamma_{h}=f_{h} \cdot g_{h}=e^{\mathbf{q}_{n} \eta} (C_{1} \sin (\sqrt{\alpha} \xi)+C_{2} \cos (\sqrt{\alpha} \xi) )
\end{align} 

\subsubsection*{Particular solution}

Also here, using separation of variables $-\gamma_{p}=f_{p}(\eta) \cdot g_{p}(\xi)$
\begin{align} 
\left(\frac{d f_{p}}{d \eta}-\frac{1}{P e^{2}} \cdot \frac{d^{2} f_{p}}{d \eta^{2}}\right) \cdot g_{p}=\frac{d^{2} g_{p}}{d \xi^{2}} \cdot f_{p}+\bar{S}_{d} \quad ; \quad \bar{S}_{d}=-\sum_{j=1}^{N} \frac{\partial \gamma_{d_{j}}}{\partial \eta} \label{3_seperation}
\end{align}
Based on the dimensionless spray BC (\ref{2_dimen_less}), we'll substitute \,{\Large $\frac{\partial \gamma_{d_j}}{\partial \eta}$}\, such that :
\begin{align} 
\bar{S}_{d}=-\sum_{j=1}^{N} \frac{\partial}{\partial \eta} \sum_{i=j}^{N} \Omega_{i j} e^{-\Delta_{i} \eta}=\sum_{j=1}^{N} \sum_{i=j}^{N} \Delta_{i} \Omega_{i j} e^{-\Delta_{i} \eta}
\end{align} 
$\Omega_{i j}(\xi) \ \Rightarrow \ \bar{S}_{d}(\xi, \eta)$ , such that $\Omega_{i j}$ can be seen as an influence coefficient on $E,$ whereas $i$ indicates the droplets influential section and $j$ indicates the influenced section. An alternative summation to these developments proposes shifting the indices such that :
\begin{align} 
\sum_{j=1}^{N} \sum_{i=j}^{N} \Delta_{i} \Omega_{i j} e^{-\Delta_{i} \eta}=\sum_{i=1}^{N} \sum_{j=1}^{i} \Delta_{i} \Omega_{i j} e^{-\Delta_{i} \eta}
\end{align} 
Such that $\bar{S}_{d}$ can be represented as :
\begin{align} 
S_{d}=\underbrace{\sum_{i=1}^{N} \Delta_{i} e^{-\Delta_{i} \eta}}_{f(\eta)} \underbrace{\sum_{j=1}^{i} \Omega_{i j}}_{g(\xi)} \quad ; \quad \sum_{j=1}^{i} \Omega_{i j} \triangleq \underbrace{H_{i}(\xi)}_{ \text {Heaviside} \atop \text { function }}= \begin{cases} 
1, & \xi>0 \\
0, & \text { else }
\end{cases} \label{3_heaviside}
\end{align} 

The latter term is dependent on the iDSD and is nullified at $H(\xi>c)=0$ . By using Fourier series we can present the Heaviside function discretely :
\begin{align} 
H_{i}(\xi)=\frac{1}{2} k_{0_{i}}+\sum_{m=1}^{\infty} k_{m_{i}} \cos (m \pi \xi) \label{3_heaviside_d}
\end{align} 
Equivalently with $\bar{S}_{d}$\, particular solution :
\begin{align} 
\gamma_{p}=\sum_{i=1}^{N}\left(\frac{1}{2} b_{0_{i}}+\sum_{n=1}^{\infty} b_{n_{i}} \cos (n \pi \xi)\right) e^{-\Delta_{i} n}
\end{align} 
After plugging inside (\ref{3_seperation}) and grouping the elements :
\begin{align} 
\sum_{i=1}^{N}\left[-\left(\Delta_{i}+\frac{\Delta_{i}^{2}}{P e^{2}}\right)\left(\frac{1}{2} b_{0_{i}}+\sum_{n=1}^{\infty} b_{n_{i}} \cos (n \pi \xi)\right)\right.&\left.+\sum_{n=1}^{\infty}(n \pi)^{2} b_{n_{i}} \cos (n \pi \xi)\right] e^{-\Delta_{i} \eta} \ldots \\
&=\sum_{i=1}^{N} \Delta_{i} e^{-\Delta_{i} \eta}\left(\frac{1}{2} k_{0_{i}}+\sum_{m=1}^{\infty} k_{m_{i}} \cos (m \pi \xi)\right) \notag
\end{align} 
Equating the coefficients to extract $b_{0_{i}}, b_{n_{i}}$ : 
\begin{align} 
\qquad \frac{1}{2} \sum_{i=1}^{N}\left(\Delta_{i}-\frac{\Delta_{i}^{2}}{P e^{2}}\right) b_{0_{i}} e^{-\Delta_{i} \eta}=\frac{1}{2} \sum_{i=1}^{N} \Delta_{i} k_{0_{i}} e^{-\Delta_{i} \eta} \\
\Rightarrow  \quad b_{0_{i}}=-\frac{\Delta_{i}}{\Delta_{i}+\left(\frac{\Delta_{i}}{P e}\right)^{2}} \cdot k_{0, i} \hspace{15mm}  \label{3_K_0i} \\
\text { Similarly with } b_{n_{i}} \quad \Rightarrow \quad b_{n_{i}}=-\frac{\Delta_{i}}{\Delta_{i}+\left(\frac{\Delta_{i}}{P e}\right)^{2}-(n \pi)^{2}} \cdot k_{n_{i}} \label{3_K_ni}
\end{align} 

Recall the \ $\gamma=\gamma_{h}+\gamma_{p}$ \ solution such that finding the coefficients of \ $C_{1}, C_{2}, \alpha$ \ can be done by matching the solution to the BC :
\begin{align} 
\left.\frac{\partial \gamma}{\partial \xi}\right|_{\xi=0}=\sqrt{\alpha} e^{\mathbf{q}_{n} \eta} C_{1}=0 & \quad \Rightarrow \quad C_{1}=0 \\
\left.\frac{\partial \gamma}{\partial \xi}\right|_{\xi=1}=-\sqrt{\alpha} e^{\mathbf{q}_{n} \eta} C_{2} \sin (\sqrt{\alpha})=0 \quad \Rightarrow \quad C_{2_{n}} \neq 0 \quad , & \quad \alpha=(n \pi)_{n \in \mathbb{N}}^{2}
\end{align}
Therefore the homogeneous solution is :
\begin{align} 
\gamma_{h}=\sum_{n=1}^{\infty} e^{\mathbf{q}_{a} \eta} C_{2_{n}} \cos (n \pi \xi) \\
\text { where } \tab  \mathbf{q}_{n} \triangleq \frac{\left(P e^{2}\right)}{2}  \bigg[1-\sqrt{1+\frac{4(n \pi)^{2}}{(P e)^{2}}} \bigg]  \label{3_qn}
\end{align} 
Applying the BC in gaseous phase equation (\ref{3_gas_phase}) :
\begin{align}
\bigg( \gamma-\frac{1}{P e^{2}} \frac{\partial \gamma}{\partial \eta} \bigg) \bigg|_{\eta=0}=\begin{cases}
1-\sum_{i=1}^{N} \delta_{i} & , \quad 0 \leq \xi \leq c \\
-V & , \quad c \leq \xi \leq 1 \end{cases}
\end{align} 

The RHS is developed using Fourier series :
\begin{align}
R H S \triangleq \frac{d_{0}}{2}+\sum_{n=1}^{\infty} d_{n_{i}} \cos (n \pi \xi) \\
d_{0}=2 \int_{0}^{c}\Big(1-\sum_{i=1}^{N} \delta_{i}\Big) d \xi+2 \int_{c}^{1}(-V) d \xi=2 c\Big(1-\sum_{i=1}^{N} \delta_{i}\Big)+2 V(c-1)
\end{align} 
Such that :
\begin{align}
d_{0}=2 c \, (1-\sum_{i=1}^{N} \delta_{i}+V )-2 V
\end{align} 
Similarly with $d_{n}$ :
\begin{align} 
d_{n}=2 \int_{0}^{c}\Big(1-\sum_{i=1}^{N} \delta_{i}\Big) \cos (n \pi \xi) d \xi-2 \int_{c}^{1} V \cos (n \pi \xi) d \xi \\
d_{n}=\frac{2\Big(1-\sum_{i=1}^{N} \delta_{i}\Big)}{n \pi} \sin (n \pi c)+\frac{2 V}{n \pi} \sin (n \pi c)
\end{align} 

Applying $\eta=0$ on the RHS :
\begin{align} 
R H S=\left(1-\sum_{i=1}^{N} \delta_{i}+V\right)\left[c+\frac{2}{\pi} \sum_{n=1}^{\infty} \frac{\sin (n \pi c)}{n} \cos (n \pi \xi)\right]-V
\end{align} 
Developing the LHS expression :
\begin{align} 
\begin{aligned}
\gamma-\frac{1}{P e^{2}} & \frac{\partial \gamma}{\partial \eta}=C_{2_{0}}+\sum_{i=1}^{N} e^{\mathbf{q}_{a} \eta} C_{2_{n}} \cos (n \pi \xi)\left(1-\frac{q_{n}}{P e^{2}}\right) \cdots \\
&+\sum_{i=1}^{N} e^{-\Delta_{i \eta}}\left(\frac{1}{2} b_{0_{i}}+\sum_{n=1}^{\infty} b_{n_{i}} \cos (n \pi \xi)\right)\left(1+\frac{\Delta_{i}}{P e^{2}}\right)
\end{aligned}
\end{align} 
Applying $LHS \Big|_{\eta=0}=$ RHS and sorting the elements :
\begin{align} \label{3_LHS_RHS}
\left.\left(\gamma-\frac{1}{P e^{2}} \frac{\partial \gamma}{\partial \eta}\right)\right|_{\eta=0}=C_{20}+\sum_{i=1}^{N} \frac{1}{2} b_{0}\left(1+\frac{\Delta_{i}}{P e^{2}}\right) \ldots \\  
+\sum_{n=1}^{\infty}\left[C_{2 n}\left(1-\frac{q_{n}}{P e^{2}}\right)+\sum_{i=1}^{N} b_{n_{i}}\left(1+\frac{\Delta_{i}}{P e^{2}}\right)\right] \cos (n \pi \xi)= \notag \\
\left(1-\sum_{i=1}^{N} \delta_{i}+V\right)\left[c+\frac{2}{\pi} \sum_{n=1}^{\infty} \frac{\sin (n \pi c)}{n} \cos (n \pi \xi)\right]-V \notag
\end{align} 
Equating the coefficients to extract $C_{2_{0}}, C_{2_{n}}$ :
\begin{align} 
C_{2_{0}}=c(1+V)-V-\sum_{i=1}^{N} c \delta_{i}+\frac{1}{2} b_{0_{i}}\left(1+\frac{\Delta_{i}}{P e^{2}}\right) \hspace{32mm} \\
C_{2_{n}}=2(1+V) \frac{\sin (n \pi c)}{n \pi\left(1-\frac{q_{n}}{P e^{2}}\right)}-\sum_{i=1}^{N} \frac{b_{n_{i}}\left(1+\frac{\Delta_{i}}{P e^{2}}\right) n \pi+2 \delta_{i} \sin (n \pi c)}{\left(1-\frac{q_{n}}{P e^{2}}\right) n \pi}
\end{align} 
Finally, after resorting the elements we can write down the full expression for $\gamma$ :
\begin{align} 
\gamma=c(1+V)-V+\frac{1}{2} \sum_{i=1}^{N}(&\left.k_{0_{i}}-2 c \delta_{i}+b_{0_{i}} e^{-\Delta_{i} \eta}\right)+\frac{2}{\pi}(1+V) \sum_{n=1}^{\infty} \frac{\sin (n \pi c)}{n\left(1-\frac{q_{n}}{P_{e}^{2}}\right)} e^{q_{n} \eta} \cos (n \pi \xi) \ldots \\
&+\sum_{n=1}^{\infty} \sum_{i=1}^{N} b_{n_{i}}\left(e^{-\Delta_{i} \eta}-\frac{\left(1+\frac{\Delta_{i}}{P e^{2}}\right) n \pi+2 \delta_{i} \sin (n \pi c)}{\left(1-\frac{q_{n}}{P e^{2}}\right) n \pi} e^{g_{n} \eta}\right) \cos (n \pi \xi) \notag
\end{align} 
Whereas the following terms $\left(k_{0}, k_{n_{i}}, b_{0_{i}}, b_{n_{i}}, q_{n}\right)$= (\ref{3_heaviside}, \ref{3_heaviside_d}, \ref{3_K_0i}, \ref{3_K_ni}, \ref{3_qn}) .

\subsection{The Temperature Equation}

Similarly to the gaseous phase development from the conservation of mass law, we can write
the temperature equation from the conservation of energy law. According to Fourier law of thermal conduction, the heat flux is linear to the spatial gradient of the temperature:
\begin{align}
q=\underbrace{-\lambda}_{   \underset{\text {conduction}}{\text {Thermal}}} \cdot
\underbrace{(\nabla \hat{T})}_{   \underset{\text {temperature}}{\text {Dimensionless}}}
\end{align}
Using the continuity equation for the energy yields the heat equation:
\begin{align}
U_{0_{g}} \frac{\partial \hat{T}}{\partial y}=K\left(\frac{\partial^{2} \hat{T}}{\partial x^{2}}+\frac{\partial^{2} \hat{T}}{\partial y^{2}}\right)+
\underbrace{\bar{S}_{\text {reaction }}}_{ \underset{\text {source}}{\text {Energy}}}+
\underbrace{ \bar{S}_{\text {d, vapor. }} }_{ \underset{\text {sink}}{\text {Energy}}}+
\underbrace{ \bar{S}_{\text {d, burning }} }_{ \underset{\text {source}}{\text {Energy}}} \quad ; \quad K=\frac{\lambda}{\rho C_{P}}
\end{align}
The $\bar{S}$ terms denote the energy transferred along the process. However, it is customary to assume that no energy transferred in the \,$\Delta \hat{T}$\, between the droplets and its carrier gas.

\subsubsection{Boundary conditions}
The BC at the channel's exit includes a thermal diffusive flux element. The temperature of both gaseous fuel and oxidizer $(=\hat{T}_{0} )$ and equals to the droplets' evaporation temperature :
\begin{align}
\hat{T}-\frac{K}{U_{0, \text { fuel }}} \frac{\partial \hat{t}}{\partial y}=\hat{T}_{0}
\end{align}
Applying the following assumptions :
\begin{align}
\underbrace{\frac{\partial \hat{T}}{\partial x}}_{\text {Symmetry }} \bigg|_{x=0} ^{y \geq 0}=0 \tab \underbrace{\frac{\partial \hat{T}}{\partial x}}_{  \underset{\text {walls}}{\text{Insulated}} } \bigg|_{x=R}^{y \geq 0}= 0 \tab \underbrace{\frac{\partial \hat{T}}{\partial y}}_{  \underset{\text {equilibrium}}{\text{Thermodynamic}} }\bigg|_{0 \leq x \leq R}^{y \rightarrow \infty}=0
\end{align}
Axes normalization is according to (\ref{2_axes_norm}) and we'll normalize the reference temperature $\left(\hat{T}_{ref}\right)$ and the injected fluid $\left(\hat{T}_{0}\right)$ temperature :
\begin{align}
(T)=\left(\frac{\hat{T}-\hat{T}_{0}}{\hat{T}_{r e f}}\right) \quad ; \quad \hat{T}_{ref}=\frac{\tilde{q}_{\text{reac}}}{m_{ \text{Tot, fuel}}\, C_{P}}
\end{align}

The energy sinks and sources elements will be normalized as such :
\begin{align}
\left(\bar{S}_{\mathrm{reac}}, \ \bar{S}_{d, v}, \ \bar{S}_{d, \mathrm{b}}\right)=\frac{R^{2}}{D_{g} \hat{T}_{ref}}\left(\tilde{S}_{\mathrm{reaction}}, \ \tilde{S}_{\mathrm{d}, \mathrm{vapor} .}, \ \tilde{S}_{\mathrm{d}, \text{ burning }}\right) \label{4_energy}
\end{align}
Such that the dimensionless temperature equation :
\begin{align}
\frac{\partial T}{\partial \eta}=L e \cdot\left(\frac{\partial^{2} T}{\partial \xi^{2}}+\frac{1}{P e^{2}} \frac{\partial^{2} T}{\partial \eta^{2}}\right)+\bar{S}_{\text{reac}}+\bar{S}_{d, v}+\bar{S}_{d, b} \label{4_temp_dless}
\end{align}
We shall work with Lewis number of :
\begin{align}
L e \triangleq \frac{K}{D_{g}}=\frac{\lambda}{\rho C_{P} D_{g}}=1
\end{align}
Meaning that the rate of the reactants' mass diffusion \underline{towards} the reaction zone, is equal to the heat's thermal diffusion from the reaction zone. Now, let us elaborate on the dimensionless sources elements. $\bar{S}_{d, v}$ expresses the heat absorption resulted from the droplets' evaporation and is equal to the evaporation rate multiplied by latent heat :
\begin{align}
\bar{S}_{d, v}=-\Lambda \, S_{d} H(c-\xi) \quad ; \quad \underbrace{H(\xi)}_{\text {Heaviside} \atop \text { function }} = \begin{cases}
1\ , & \xi>0 \\
0\ , & \text { else }
\end{cases}
\end{align}

Note that the Heaviside step function acts as an "on / off" switch controlling the source. $\bar{S}_{d, b}$ expresses the external heat emitted from the moving burned droplets, after being ignited. Using former assumptions we get the equality $E_{\text {burned}}=E_{\text {pre-ignited}}$ such that :
\begin{align}
\bar{S}_{d, b}=\bar{S}_{d} H(c-\xi) H\left(\eta-\eta_{f}\right)
\end{align}
$\eta_{f}$ denotes the flame height with respect to the \,$\xi$ axis and $H(\xi)$ indicates the obtained heat regions. Using S-Z transform we'll define :
\begin{align}
\gamma_{T} \triangleq T+\gamma_{F} \label{4_gamma_T}
\end{align}
Such that by utilizing Eqs. (\ref{2_axes_norm}, \ref{4_temp_dless}) we get :
\begin{align}
\frac{\partial \gamma_{T}}{\partial \eta}=\frac{\partial^{2} \gamma_{T}}{\partial \xi^{2}}+\frac{1}{P e^{2}} \cdot \frac{\partial^{2} \gamma_{T}}{\partial \eta^{2}}+(1-\Lambda) \bar{S}_{d} H(c-\xi) 
\end{align}

The dimensionless BC for the temperature equation :
\begin{align}
T=T_{0}+\underbrace{\left(\frac{\lambda}{\rho C_{P} D_{g}}\right)}_{L e=1}\left(\frac{D_{g}^{2}}{U_{0_{g}}^{2} R^{2}}\right) \frac{\partial T}{\partial \eta} \quad \underset{\eta=0}{\Rightarrow} \quad T-\frac{1}{P e^{2}} \frac{\partial T}{\partial \eta}=T_{0}
\end{align}
And on the normalized axes
\begin{align}
\left.\frac{\partial T}{\partial \xi}\right|_{\xi=0,1} ^{\eta \geq 0}=\left.0 \quad \frac{\partial T}{\partial \eta}\right|_{0 \leq \xi \leq 1} ^{\eta \rightarrow \infty}=0
\end{align}
The dimensionless BC for the gaseous fuel :
\begin{align}
\gamma_{F}-\frac{1}{P e^{2}} \frac{\partial \gamma_{F}}{\partial \eta}=\left\{\begin{array}{ll}
1-\sum_{i=1}^{N} \delta_{i} & , \quad 0 \leq \xi \leq c \\
0 & , \quad c \leq \xi \leq 1
\end{array}\right.
\end{align}
Its normalized axes satisfy
\begin{align}
\left.\frac{\partial \gamma_{F}}{\partial \xi}\right|_{\xi=0,1}^{\eta \geq 0}=\left.0 \tab\frac{\partial \gamma_{F}}{\partial \eta}\right|_{0 \leq \xi \leq 1} ^{\eta \rightarrow \infty}=0
\end{align}
Finally we get the full expression for S-Z transform (\ref{4_gamma_T}) such that dimensionless $\gamma_{T}$ :
\begin{align}
\gamma_{T}-\frac{1}{P e^{2}} \frac{\partial \gamma_{T}}{\partial \eta}=\left\{\begin{array}{ll}
1-\sum_{i=1}^{N} \delta_{i}+T_{0} & , \quad 0 \leq \xi \leq c \\
T_{0} & , \quad c \leq \xi \leq 1
\end{array}\right. \label{4_SZ_trans}
\end{align}
Whose normalized axes satisfy :
\begin{align}
\left.\frac{\partial \gamma_{T}}{\partial \xi}\right|_{\xi=0,1} ^{\eta \geq 0}=\left.0 \tab \frac{\partial \gamma_{T}}{\partial \eta}\right|_{0 \leq \xi \leq 1} ^{\eta \rightarrow \infty}=0 \label{4_gamma_axes}
\end{align}

\subsubsection{Analytical solution}

Recall $\gamma_{T} \triangleq T+\gamma_{F}(4.11)$ and the dimensionless temperature equation (4.12)
\begin{align}
\frac{\partial \gamma_{T}}{\partial \eta}=\frac{\partial^{2} \gamma_{T}}{\partial \xi^{2}}+\frac{1}{P e} \cdot \frac{\partial^{2} \gamma_{T}}{\partial \eta^{2}}+(1-\Lambda) S_{d} H(c-\xi)
\end{align}
Using the BC elaborated at (\ref{4_SZ_trans}, \ref{4_gamma_axes}) we can solve $\gamma_{T}$ as a sum of $\gamma_{T}=\gamma_{T_{h}}+\gamma_{T_{p}}$
\begin{align}
\text { Homogenous : } \frac{\partial \gamma_{T_{h}}}{\partial \eta}=\frac{\partial^{2} \gamma_{T_{h}}}{\partial \xi^{2}}+\frac{1}{P e^{2}} \cdot \frac{\partial^{2} \gamma_{T_{h}}}{\partial \eta^{2}} \hspace{13mm} \\
\text { Particular : } \frac{\partial \gamma_{T_{p}}}{\partial \eta}=\frac{\partial^{2} \gamma_{T_{p}}}{\partial \xi^{2}}+\frac{1}{P e^{2}} \cdot \frac{\partial^{2} \gamma_{T_{p}}}{\partial \eta^{2}}+(1-\Lambda) \bar{S}_{d}
\end{align}

\subsubsection*{Homogeneous solution}

Using the same solution method as in the gaseous phase section
\begin{align}
\gamma_{T_{h}}=f_{T_{h}} \cdot g_{T_{h}}=e^{\mathbf{q}_{n} \eta}\left(C_{1} \sin (\sqrt{\alpha} \xi)+C_{2} \cos (\sqrt{\alpha} \xi)\right)
\end{align}

\subsubsection*{Particular solution}

We'll define the following variable, and plug terms from (\ref{3_heaviside}) :
\begin{align}
\bar{S}_{d}^{\prime}=(1-\Lambda) \bar{S}_{d}=(1-\Lambda) \underbrace{\sum_{i=1}^{N} \Delta_{i} e^{-\Delta_i \eta}}_{f(\eta)} \underbrace{\sum_{j=1}^{i} \Omega_{i j}}_{g(\xi)}
\end{align}
Also here, we'll define:
\begin{align}
H_{i}^{\prime}(\xi) \triangleq(1-\Lambda) H_{i}(\xi)=(1-\Lambda) \sum_{j=1}^{i} \Omega_{i j} \\
\text{Using Fourier} \ \Rightarrow \ H_{i}^{\prime}(\xi)=\frac{1}{2} k_{0_{i}}^{\prime}+\sum_{m=1}^{\infty} k_{m_{i}}^{\prime} \cos (m \pi \xi)
\end{align}

Its coefficients are dependent on the following relations (\ref{3_heaviside}, \ref{3_heaviside_d}) :
\begin{align} \label{4_kp_i}
\left(k_{0_{i}}^{\prime}, k_{m_{i}}^{\prime}\right)=(1-\Lambda)\left(k_{0_{i}}, k_{m_{i}}\right)
\end{align}
Such that the homogeneous equation solution is ( $q_{n} \text { @ } $ \ref{3_qn} ) :
\begin{align}
\gamma_{T_{p}}=\sum_{i=1}^{N}\left(\frac{1}{2} b_{0_{i}}^{\prime}+\sum_{n=1}^{\infty} b_{n_{i}}^{\prime} \cos (n \pi \xi)\right) e^{\mathbf{q}_{n} \eta}
\end{align}
Using similar relations as before (s.t. $\eta=0$ ):
\begin{align}
\Rightarrow \quad b_{0_{i}}^{\prime} =-\frac{\Delta_{i}}{\Delta_{i}+\left(\frac{\Delta_{i}}{P e}\right)^{2}} \cdot k_{0, i}^{\prime} \label{bp_i} \hspace{16.5mm} \\
\text { Similarly with } b_{n_{i}}^{\prime} \quad \Rightarrow \quad b_{n_{i}}^{\prime}=-\frac{\Delta_{i}}{\Delta_{i}+\left(\frac{\Delta_{i}}{P_{e}}\right)^{2}-(n \pi)^{2}} \cdot k_{n_{i}}^{\prime} \label{bp_n}
\end{align}
Applying $(V=0, \eta=0)$ on the RHS :
\begin{align}
R H S\Big|_{\eta=0}=c\left(1-\sum_{i=1}^{N} \delta_{i}+T_{0}\right)+\frac{2}{\pi}\left(1-\sum_{i=1}^{N} \delta_{i}\right) \sum_{n=1}^{\infty} \frac{\sin (n \pi c)}{n} \cos (n \pi \xi)
\end{align}
Based on (\ref{3_LHS_RHS}) development we'll sort the elements:
\begin{align}
\left.\left(\gamma-\frac{1}{P e^{2}} \frac{\partial \gamma}{\partial \eta}\right)\right|_{\eta=0}=C_{2_{0}}^{\prime}+\sum_{i=1}^{N} \frac{1}{2} b_{0_{i}}\left(1+\frac{\Delta_{i}}{P e^{2}}\right) \ldots \\
+\sum_{n=1}^{\infty}\left[C_{2_{n}}^{\prime}\left(1-\frac{q_{n}}{P e^{2}}\right)+\sum_{i=1}^{N} b_{n_{i}}^{\prime}\left(1+\frac{\Delta_{i}}{P e^{2}}\right)\right] \cos (n \pi \xi)
\end{align}
Equating the coefficients to extract $C_{2_{0}}^{\prime}, C_{2_{n}}^{\prime}$
\begin{align}
C_{2_{0}}= c+T_{0}-\sum_{i=1}^{N} c \delta_{i}+\frac{1}{2} b_{0_{i}}^{\prime}\left(1+\frac{\Delta_{i}}{P e^{2}}\right) \hspace{29mm} \\
C_{2_{0}}^{\prime} =\frac{2 \sin (n \pi c)}{n \pi\left(1-\frac{q_{n}}{P e^{2}}\right)}-\sum_{i=1}^{N} \frac{b_{n_{i}}^{\prime} \left(1+\frac{\Delta_{i}}{P e^{2}}\right) n \pi+2 \delta_{i} \sin (n \pi c)}{\left(1-\frac{q_{n}}{P e^{2}}\right) n \pi}
\end{align}

Finally, after resorting the elements we can write down the full expression for $\gamma_{T}$ :
\begin{align}
\gamma_{T}=c+T_{0}+& \frac{1}{2} \sum_{i=1}^{N}\left(k_{0_{i}}^{\prime}-2 c \delta_{i}+b_{0_{i}}^{\prime} e^{-\Delta_{i} \eta}\right)+\frac{2}{\pi} \sum_{n=1}^{\infty} \frac{\sin (n \pi c)}{n\left(1-\frac{q_{n}}{P e^{2}}\right)} e^{\textbf{q}_n \eta} \cos (n \pi \xi) \ldots \\
&+\sum_{n=1}^{\infty} \sum_{i=1}^{N} b_{n_{i}}^{\prime}\left(e^{-\Delta_{i} \eta}-\frac{\left(1+\frac{\Delta_{i}}{Pe^{2}}\right) n \pi+2 \delta_{i} \sin (n \pi c)}{\left(1-\frac{\textbf{q}_n}{Pe^{2}}\right) n \pi} e^{\textbf{q}_n \eta}\right) \cos (n \pi \xi)
\end{align}
Whereas the following terms $\left(k_{0_{0}}^{\prime}, k_{n_{i}}^{\prime}, b_{0_{i}}^{\prime}, b_{n_{i}}^{\prime}, q_{n}\right)=$ (\ref{4_kp_i}, \ref{4_kp_i}, \ref{bp_i},  \ref{bp_n}, \ref{3_qn}) .

\newpage

\subsection{Results}

Based on (\ref{4_gamma_T}) the dimensionless temperature field can be expressed as : 
\begin{align}
T(\xi, \eta) = \begin{cases}
\gamma_{T}-\gamma \tab \textbf{if} \tab \xi \leq \xi(\gamma \approx 0) \\
\gamma_{T} \hspace{17.5mm} \textbf{if} \tab \xi(\gamma \approx 0)<\xi \leq R \end{cases}
\end{align}
The flame front contains the set of points that fulfill the stoichiometric ratio between the reactants and thus lies on the black dashed line \,$\xi(\gamma \approx 0)$ . In the gaseous fuel zone the mass fraction satisfies - $\gamma = \gamma_{F}$, where in the oxidizer zone it's nullified - $\gamma_{F}=0$ : 
\begin{figure}[H]
\centering
\includegraphics[width=0.97\linewidth, center]{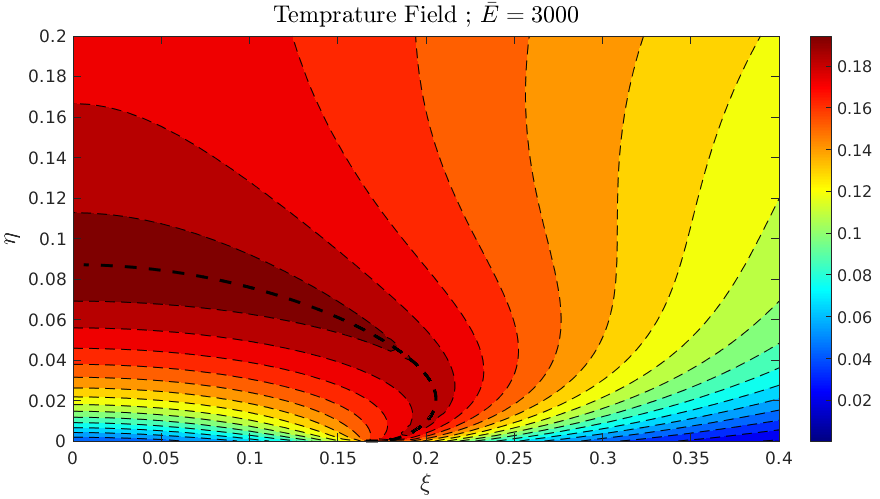}
\end{figure}
We can see that the flame base is formed at half the inner channel \,$c = \frac{L}{R}$\,, while its maximum height is obtained at the middle of the symmetric flame. Trivially, different input setups would result in different values of the temperature field. 

However, of all existing points, we are interested in two specific points  that might shed light on the performance of a given execution, obtained via the following functions : 

\tab $\circ$ \ $\eta_{max} \,= f_{\eta}(\bar{E}, d, \delta) \ \Rightarrow \ $ Returns the highest value of the flame front.

\tab $\circ$ \ $T_{max} = f_{T}(\bar{E}, d, \delta) \ \Rightarrow \ $ Returns the highest value of the temperature field.

\newpage

The temperature field can be also seen as a set of \hypertarget{field_3D}{3D points} - $\big(\xi, \, \eta, \, T(\xi, \eta) \big)$ : 
\begin{figure}[H] 
\centering
\includegraphics[width=1.0\linewidth, center]{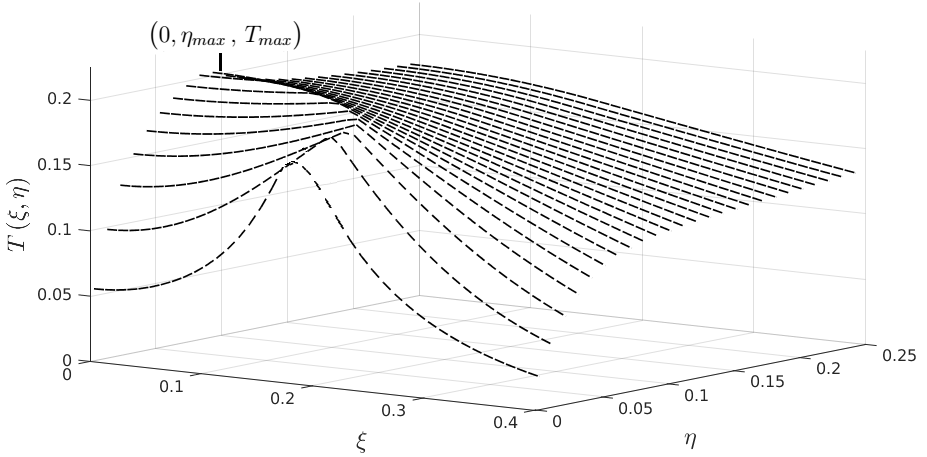}
\end{figure} 

Both variables at the edge of the front flame will serve as the main indicators of a given execution. In the following subsections we will further inspect them, in order to establish a general intuition with respect to several different factors.

\newpage

\subsubsection{Maximum flame height}

The following simulations examine the model’s sensitivity in terms of maximum flame height, namely \,$\eta_{max}$\, reaction to several initial parameters : (i) liquid fuel fraction - $\delta_i$ (ii) the droplets section - $d$ and (iii) the evaporation rate - $\bar{E}$. Different reactions to the iDSD composition were observed, and thus the following scenario will present them.

\subsubsection*{Monosectional iDSD}
An iDSD is said to be monosectional when the liquid fuel fraction occupies only one section, namely the initial droplet size is uniform. Consider the following set of monosectionals, executed at a wide range of sections and varying amounts of liquid fuel :
\begin{figure}[H]
\centering
\includegraphics[width=1.175\linewidth, center]{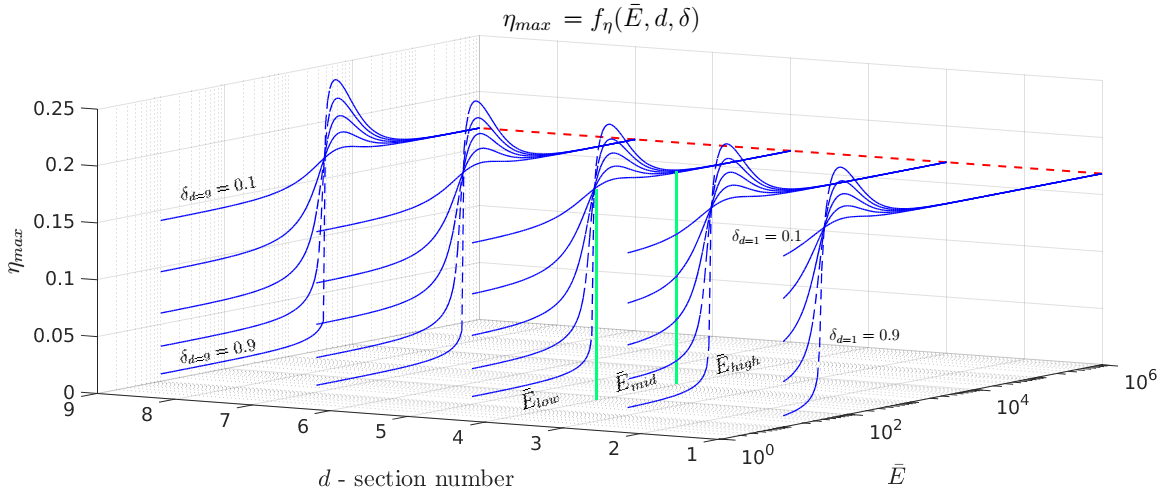}
\end{figure}
Given five section sizes - $d = \{ 1:2:9 \}$ , each contains five executions of different amounts of liquid fuel - $\delta_d = \{ 0.1:0.2:0.9 \}$ . Note the overall tendency of \,$\eta_{max}$\, to increase along the section sizes, in a typical tradeoff with the evaporation rate $\bar{E}$. 

The red dashed line denotes the gaseous flame height towards highest $\bar{E}_{\rightarrow \infty}$, see detailed discussion next.

\subsubsection*{Polysectional iDSD} \label{rand_init}
The polysectional scenario is an iDSD characterized by a multimodal distribution : 
\begin{figure}[H]
\centering
\includegraphics[width=0.745\linewidth, center]{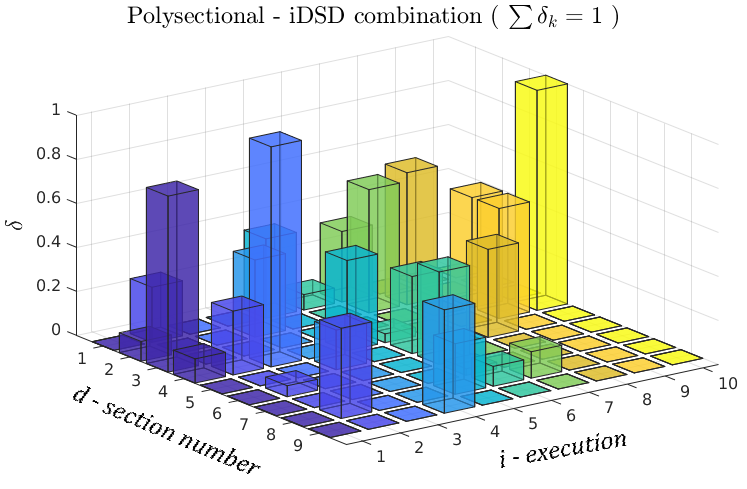}
\end{figure}
In this scenario, executions $i= [1 \, , \, 9]$ are polysectionals composed of random iDSD initialization, whereas each total sum equals to one. Execution $i=10$ acts as a "control group" as it has only one section, namely a monosectional of $\delta_{d=4} = 1.0$ .
\begin{figure}[H]
\centering
\includegraphics[width=1.15\linewidth, center]{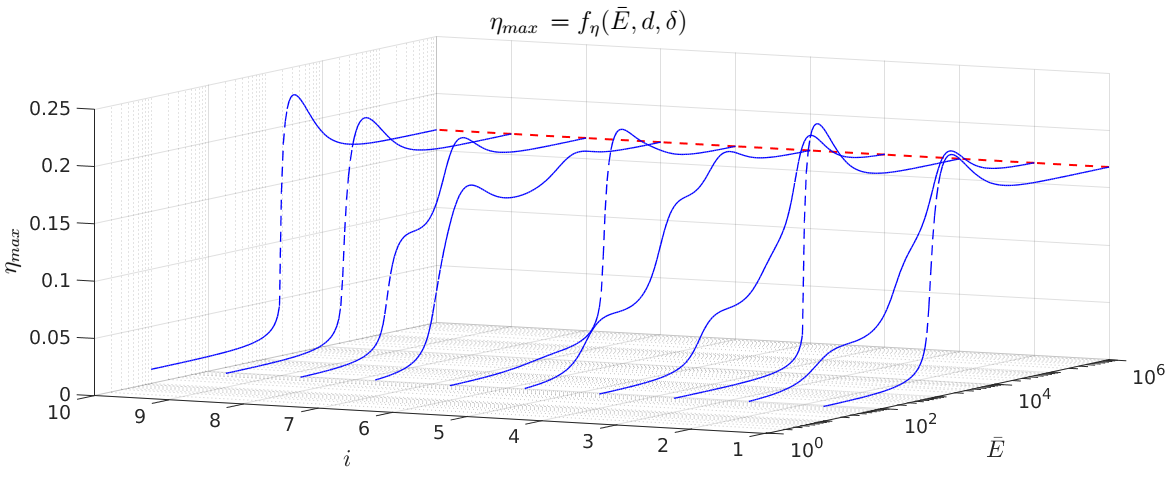}
\end{figure}
Note the typical differences between the \underline{monosectional} iDSD and the \underline{polysectional}.

We can see an interesting pattern that characterizes all of the cases :
\begin{figure}[H] 
\centering
\includegraphics[width=0.975 \linewidth, center]{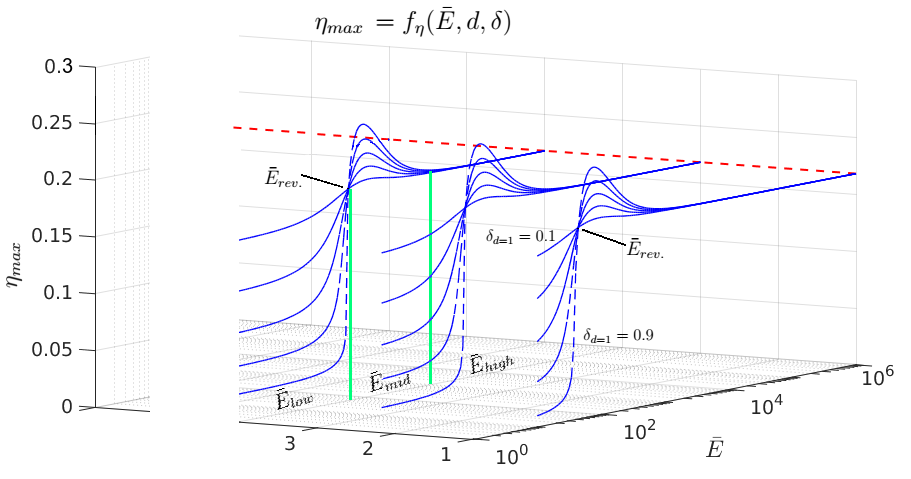}
\end{figure}
By drawing three different \underline{qualitative} $\bar{E}$ zones (vertical green separation) : 
\begin{align}
\text{\hspace{2.5mm}$\bar{E}_{low}$ := \ $\eta_{max}$\, is governed mainly by the amount of liquid fraction ($\delta$) } \label{E_low} \\
\text{$\bar{E}_{mid}$ \ := \ $\eta_{max}$\, is dependent significantly by the evaporation rate \,($\bar{E}$) } \hspace{3mm}  \label{E_mid} \\
\text{$\bar{E}_{high}$ := \,$\eta_{max}$\, approaches to the gaseous flame height (dashed red) : } \hspace{2.5mm}  \label{E_high} \\
\lim_{\bar{E}\rightarrow \infty} f\, (\bar{E}) \approx 0.17 \quad \forall \quad \{ d, \, \delta \} \hspace{35mm} \notag
\end{align}

Additionally, let us define the intersection between $\bar{E}_{low}$ and $\bar{E}_{mid}$ as a \textit{reversal} point - $\bar{E}_{rev.}$ from which bigger fuel fractions yield higher flame, as opposed to before \,($\bar{E} < \bar{E}_{rev.}$) . 

Note the $\{ d \, - \, \bar{E} \}$ correlation, and its impact on the flame height : \\
\quad $\circ$ \ Small particles ($d=1$) yield optimal flame at relatively low \, $\bar{E}_{mid} \approx 100 \hspace{3mm} \forall \hspace{3mm} \delta$. \\
\quad $\circ$ \ Larger particles ($d=5$) yield optimal flame relatively higher $\bar{E}_{mid}  \approx 1000 \hspace{3mm} \forall \hspace{3mm} \delta$. 


\subsubsection{Maximum tip flame temperature}
In this scenario I would like to go through the same process regarding \,$T_{max}$ .

\subsubsection*{Monosectional iDSD}
Starting off with the same setup of monosectional iDSD executed previously :
\begin{figure}[H] 
\centering
\includegraphics[width=1.1 \linewidth, center]{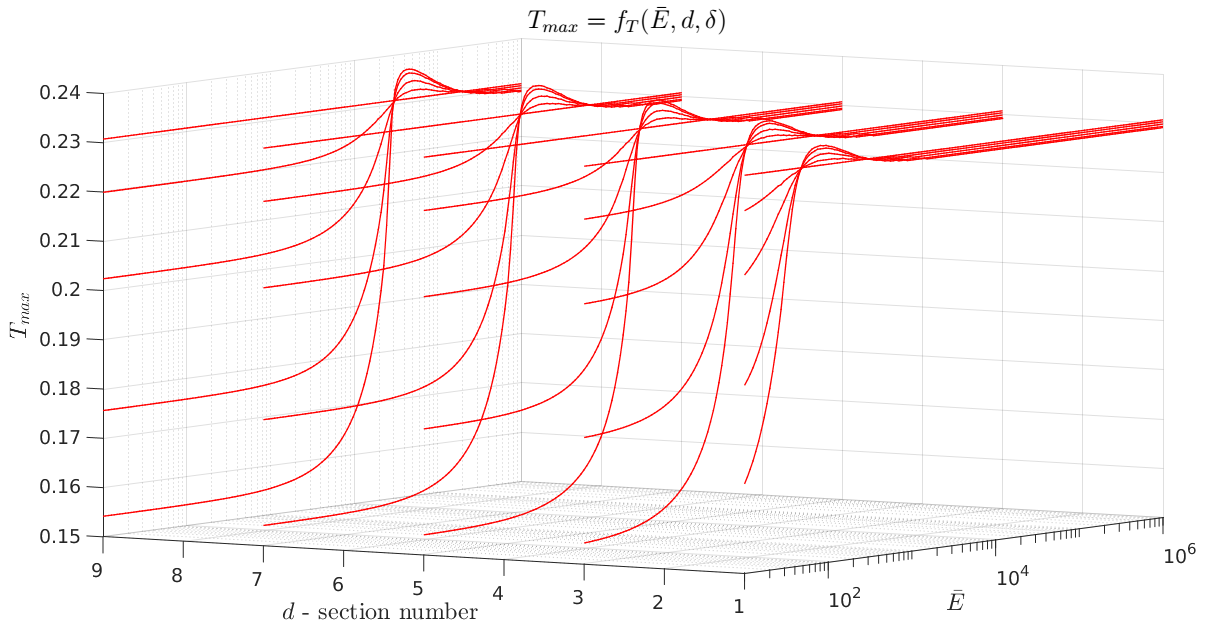}
\end{figure}

In a closer look around the optimal zone $\bar{E}_{mid}$ :
\begin{figure}[H] 
\centering
\includegraphics[width=1.1 \linewidth, center]{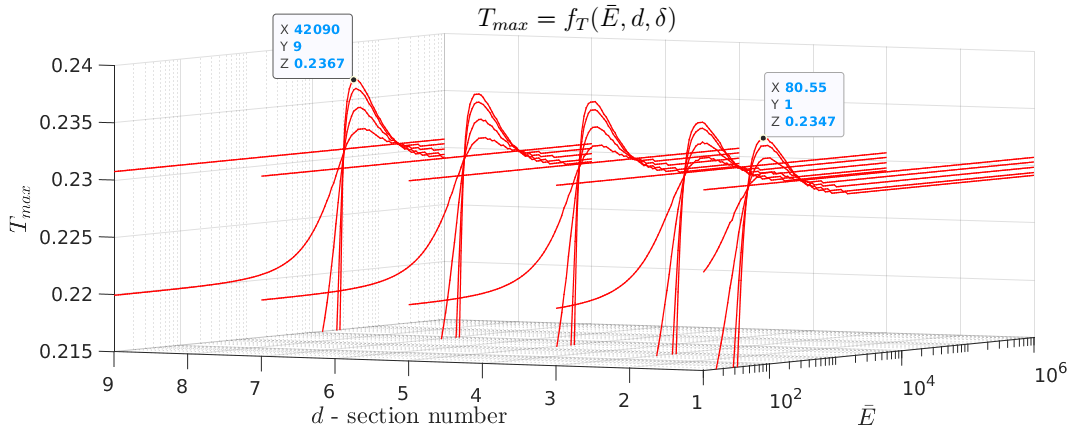}
\end{figure}
Either here, in the absence of any constraints, we get $T_{max} \, \uparrow$\, for any $\{ \, \delta \uparrow \ , \ d \uparrow \,  \}$ .

\subsubsection*{Polysectional iDSD}
Using a random initialization for all executions except $i=6$, which is a monosectional :
\begin{figure}[H] 
\centering
\includegraphics[width=1.04 \linewidth, center]{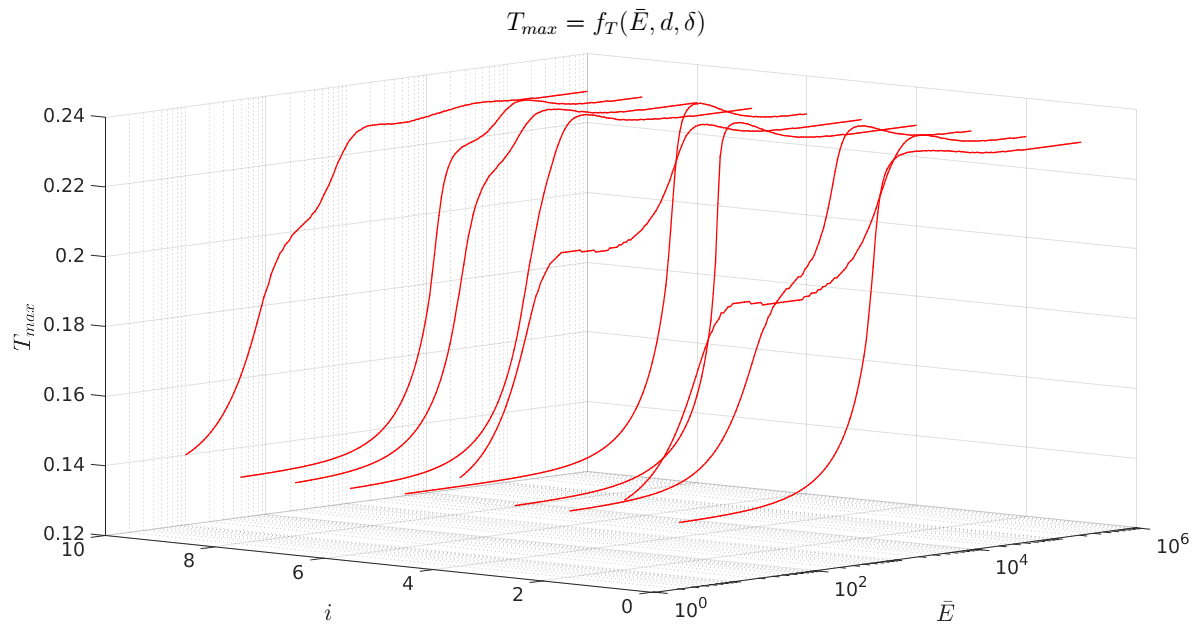}
\end{figure}
Mind the graphs curvatures that later will be discussed. At a closer look around $\bar{E}_{mid}$ :
\begin{figure}[H] 
\centering
\includegraphics[width=1.04 \linewidth, center]{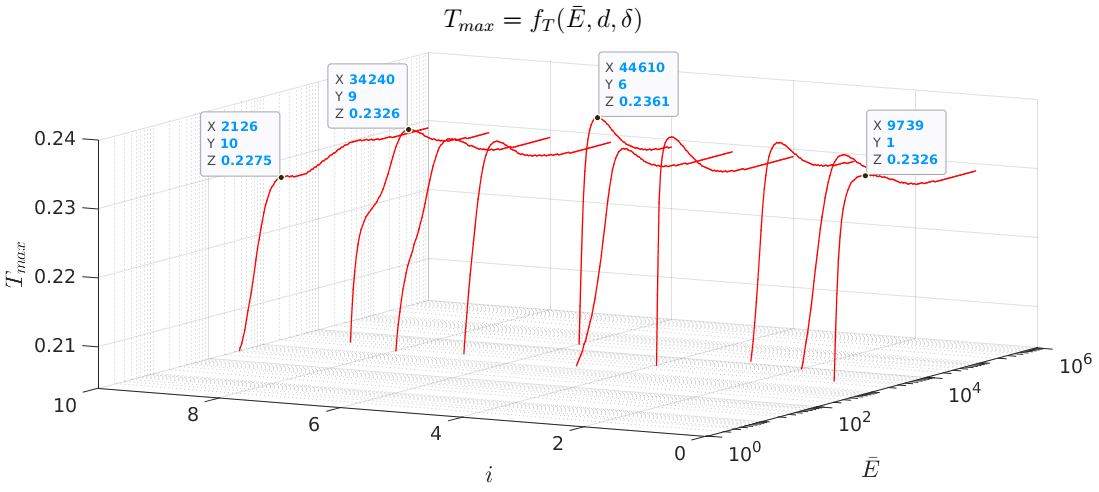}
\end{figure}
Similarly to what we saw at \,$\eta_{max}$, the "hottest" $T_{max}$ is obtained for $i=6$, which is the only sample that \underline{is not} a polysectional iDSD.

\subsection{Discussion}

Both $\eta_{max}$ and $T_{max}$ exhibited a typical pattern as elaborated on (\ref{E_low} - \ref{E_high}). At both cases, the polysectional iDSD performed poorer in comparison with the monosectional. Towards higher evaporation rates ( $\bar{E}_{high}$ ), it seemed that all of the graphs approach the gaseous flame \underline{height}, regardless their initial conditions. Here - $\eta_{max}  \approx 0.17 \hspace{2mm} \forall \hspace{2mm} \{ \, d \ , \delta \ \} $ . 
\begin{figure}[H] 
\centering
\includegraphics[width=0.6 \linewidth, center]{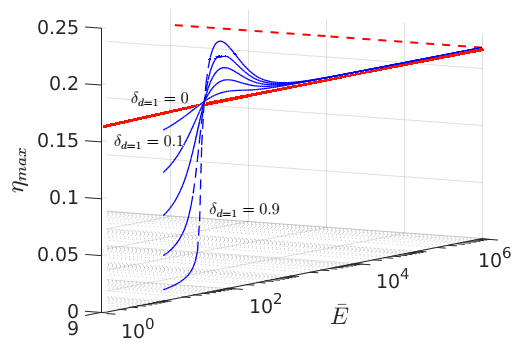}
\end{figure}

Quite similarly, the temperature graphs show an asymptotic behavior towards \,$\bar{E}_{high}$ . The gaseous flame \underline{temperature} satisfies \,$T_f \approx 0.2307$ , but performed somewhat oppositely at higher $\bar{E}$, where \,$T_{max}$ reaction's benefits with \underline{smaller} amounts of the liquid fuel :
\begin{figure}[H] 
\centering
\includegraphics[width=0.8 \linewidth, center]{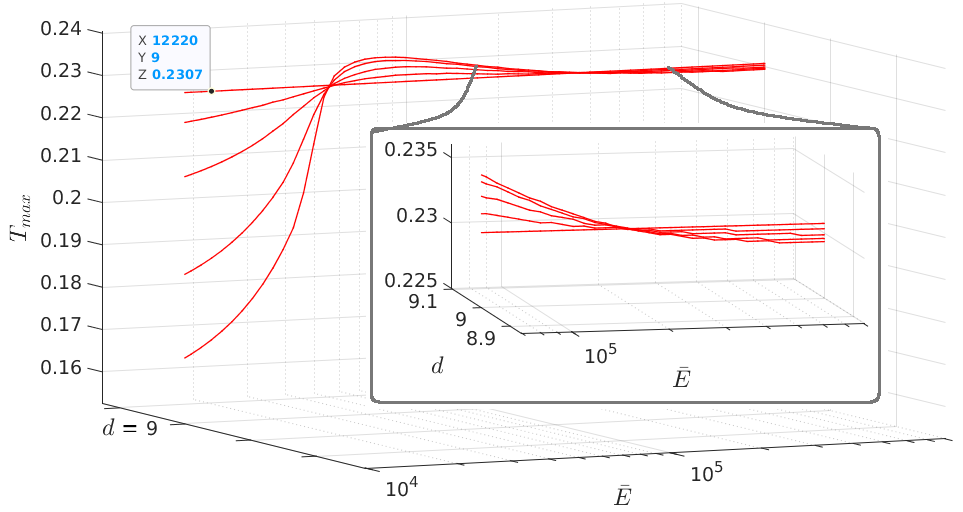}
\end{figure}

Looking closely, we can tell that slow evaporation rates benefit with small $\delta$ amounts contrarily to bigger  $\delta$ that perform significantly poorer. That mechanism can be explained by the liquid ”overload” that absorbs more heat during the evaporation process : 
\begin{figure}[H] 
\centering
\includegraphics[width=1.0 \linewidth, center]{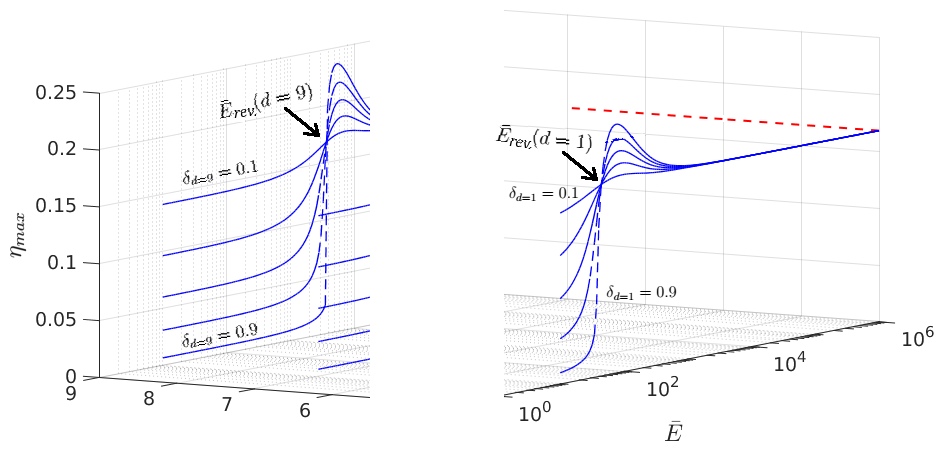}
\end{figure}

The qualitative nature of the graphs shows how the \,$\bar{E}_{rev.}$\, changes with respect to $d$ section and acts as a ”role reversal” between $\delta$ and $\bar{E}$ . Note that the evaporation rate axis is \underline{logarithmic}, and hence the differences turn more dramatic as we move forward.

The polysectional iDSD have shown \underline{poorer} performances in comparison with the monosectional, as it affects directly the droplets volume change : \\
\quad (i) \hspace{1mm}It downsizes the droplets size inside a \underline{given} section. \\
\quad (ii) It is responsible for the joining of new droplets from \underline{higher} sections.

It is therefore no surprise that either \,$\eta_{max}$ or $T_{max}$\, curvatures, are appearing in accordance with occupied sections.

\newpage

\section{The Optimization Model}
In this section the combustion model will be adjusted to the optimization model. At first, a formal development of the GA will be presented, then the combustion model will be merged within. Consider the following informal description that demonstrates that process : \vspace{.2cm}

\begin{algorithm}[H] 
	\SetAlgoLined
	Initialize random population \ $ \Big( X_t \in \mathbb{R}^{n \times 3} \Big)$ : \\	
	$t \leftarrow 0$ \\
	\For{$x_j \ \in \ X_{0} \ $}{
			\quad $x_j \leftarrow  random \, \big(\, \bar{E}, \, d, \, \delta \, \big) \ \in \ \mathbb{R}^3$ \\}
	
	\BlankLine
	Commence cost minimization until termination condition is met : \\
	\While{ $t \ < \ GA_{\, MaxIteration}$ \ or \ $GA_{\, Cost} \ > \ GA_{\, Termination} \ $ }{	
		\BlankLine
		\text{Evaluate Fitness():} \\
		\tab $Y_t \leftarrow  f \,(X_t)$ \\

		Selection():\\
		\tab $[\, x^{I}_t\, , \, x^{II}_t \,] \leftarrow X_t \Big( \max(\, Y_t \, )_{k=2} \Big)$ \\
		Crossover($X_t$): \\
		\tab $[\, x^{\prime}_t\, , \, x^{\prime \prime}_t \,] \leftarrow recombine \Big( \, x^{I}_t\, , \, x^{II}_t \, \Big)$ \\
		Mutation(): \\
		\hspace{11mm} $X_{t+1} \leftarrow$ $Mutate \Big( \, x^{\prime}_t\, , \, x^{\prime \prime}_t \, \Big)$ \\		
		$t \leftarrow t+1$
	}
	\BlankLine
	
Extract best candidate from last generation : \\
\textbf{Return} \, $x_j^* \leftarrow X_t \Big( \max(\, Y_t \, )_{k=1} \Big) $
\caption{Pseudocode of the Genetic Algorithm}
\end{algorithm}  \vspace{.1cm}

\textbf{Legend} 

%

\begin{tabular}{ p{2.5 cm} p{12cm}}
$f  \,( \, \cdot \, )$ & Fitness function ( $f_{\eta}$ or $f_T$ ) \\
$x_t$ & Chromosome at time $t$ \\
$y_t$ & Chromosome's fitness at time $t$ \\
$X_t$ & Population at time $t$ \\
$Y_t$ & Population's fitness at time $t$ \\
$\max(\, Y_t \, )_{k}$ & Find $k$ \underline{largest} elements \\
$X_t \big(\, g( \, \cdot \, ) \, \big)$ & Extract \,$x_t \in X_t$\, that satisfies $ g( \, \cdot \, )$ \\
\end{tabular} 

\textbf{Note :} The\, $GA_{text}$\, denotes the GA's internal functions which are problem-independent, and are decoded in accordance with the algorithmic design. Moreover, the GA uses a \underline{bit-string} representation to encode the solutions, in the form of chromosomes.

\subsection{Problem formulation}
The canonical form of an optimization problem [\ref{c1}] is commonly written as :
\begin{align}
&\underset{\mathbf{x}}{\operatorname{minimize}} & f(\mathbf{x}) \label{f_opt}\hspace{35.5mm} \\
&\operatorname{subject \ to}
& g_i(\mathbf{x}) \leq 0, \hspace{3mm} i = 1, \dots, m  \label{Inequality} \\
&& h_i(\mathbf{x}) = 0, \quad i = 1, \dots, p  \label{Equality} 
\end{align}
where
\begin{align*}
f\, : \, \mathbb{R}^n \rightarrow \mathbb{R} & & \text{The objective function} \\
g_i(x) \leq 0 & & \text{Inequality constraint} \\
h_j(x) = 0 & & \text{Equality constraint}
\end{align*}
As mentioned in Definition [ \hyperlink{convex_concave}{B.3} ], a \textbf{maximization} problem can be solved analogously by simply negating the objective function : \,$\max_x f(x) \, \Leftrightarrow \, \min_x -f(x) $ . 

The first technical obstacle I encountered, was Matlab's inability to execute equal constraints while one of the search variables is \underline{discontinuous}. The GA becomes inoperable, as the section variable (\,$d$\,) is restricted to be an integer \,$\mathbb{N}_{>0}$ , and thus breaks the continuity assumption of the search space [\ref{continuity}]. Commonly, these class of problems can be addressed as "Mixed-integer linear programming" (MILP) [\ref{MILP_1}],[\ref{MILP_2}]. 

However, after enough search the answer was found in one of Matlab's forums [\ref{matlab_problem}], that proposed to achieve the equality constraint by setting two inequality constraint :
\begin{align}
\text{desired :} \quad a x_1 + b x_2 = c \tab \text{(impossible)} \label{impossible} \hspace{10.5mm} \\
\text{workaround :} \label{workaround}
\quad a x_1 + b x_2 \leq c \quad \& \ \, - a x_1 - b x_2 \leq - c
\end{align}
Practically, the executions were conducted as in Eq. (\ref{workaround}), but for simplicity reasons I will write them regularly as an equality constraint, as in Eq. (\ref{impossible}).

As mentioned above, there are two types of problem's constraints : \\
\hspace{15mm} (\ref{Inequality}) - Inequality constraint \hspace{15mm}
(\ref{Equality}) - Equality constraint

While inequality imposes a half-constraint on a variable between a certain point onwards, the equality constraint imposes a full-constraint. In other words, an equality constraint on a single variable, degenerates its participation in the optimization process, and thus \underline{reduces} the overall degree of freedom (DoF) by one. Contrarily, a variable that is not subjected to any constraint is said to be \underline{unconstrained}, as it is completely free to be sampled within the domain \,$x_k \in (-\infty , \infty)$, and thus \underline{contributing} 1 DoF to the system.

\subsection{Explicit formulation}
Here, the objective functions can return either \,$\eta_{max}$\, or \,$T_{max}$ . The inequality constraint is the liquid fraction that's constrained to be smaller than a desired value, and the equality constraint is the constant evaporation rate. On \textit{Matlab} : 
\begin{figure}[H] 
\centering
\includegraphics[width=0.875 \linewidth, center]{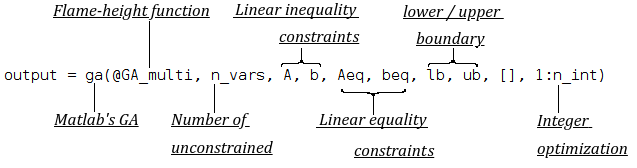}
\end{figure}
Thereby, the program is free to find the optimal iDSD as \,$d$\, is an unconstrained variable :
\begin{align*}
& \hspace{11mm} \underset{ \bar{d} }{\operatorname{max}} & f( \bar{E}, \, d, \, \delta ) \hspace{43mm} \\
&\operatorname{subject \ to}
& 1 \leq d_i \leq 9  \quad \forall \quad i = 1, \, ..., \, n_{sections}  \hspace{1mm} \\
&& \sum_i \delta_i \leq \delta_{const} \leq 1 \quad ; \quad \bar{E} = \bar{E}_{const}
\end{align*}

The $i$ index denotes the \underline{serial} section number, which cannot be greater than the total number of sections ( $i \leq 9$ ). The algorithm is completely free to optimize the search space over \underline{any} possible combination of sections / iDSD, such that eventually each one of the sections corresponds to its liquid fraction ( $ d_i \ \widehat{=} \ \delta_i$ ). 

Consider the following 3 DoF toy-example, which optimizes the iDSD$\big|_{i \leq 3}$  : 
\begin{figure}[H] 
\centering
\includegraphics[width=0.75 \linewidth, center]{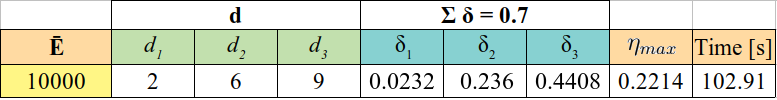}
\end{figure}
We can see that out of all feasible combinations, the optimal distribution consists of the 2nd, 6th and 9th sections, each correlates to a different index of liquid fraction. 

The following figure shows the on-line optimization process as presented by \textit{Matlab} : \vspace{0.5cm}

\begin{figure}[H]
\centering
\includegraphics[width=1.0 \linewidth, center]{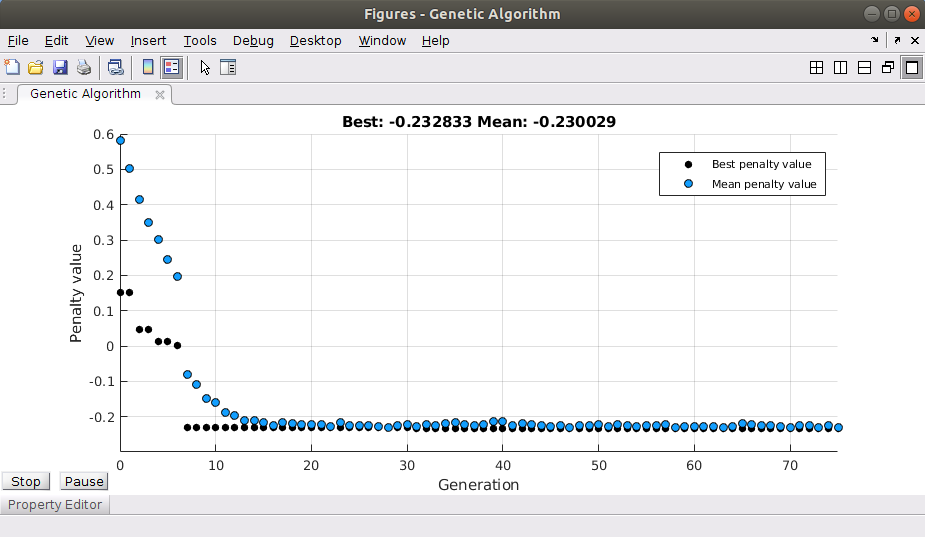}
\end{figure}

The x-axis denotes the iterative process of forming better generations that can better perform the fitness function on the y-axis (penalty value). Note that the values here are expressed as negative, as by default the algorithmic is designed to minimize a penalty. Therefore the problem was equivalently adjusted, by changing the output sign [\,\hyperlink{convex_concave}{B.3}\,] .
\newpage


\section{Results}
This section presents a wide set of examinations of the optimal flame height and the optimal tip flame temperature. Each examination is conducted at growing levels of DoF, over several evaporation rates, starting from a single DoF towards 9 DoF (total sections). 

However, in order to fit with the algorithm requirements, it must fulfill two constraints : 

\quad (i) \ $\sum_i \delta_i \leq 0.7$ - the sum of liquid fuel can be at most 0.7 (at all sections) 

\quad (ii) $\bar{E}=\text{const.}$ \,- the evaporation rate must be constant with time

Thereby, the algorithm will be free to compute the optimal iDSD in return. 

\textbf{Note} : $d$ (number of sections) and DoF (degree of freedom) are used interchangeably.

\vspace{3cm}

\subsection{Flame height optimization}
Starting from a single section optimization, we'll utilize \,$\sum_i \delta_i \leq 0.7$\, as a study case :
\begin{figure}[H] 
\centering
\includegraphics[width=0.34 \linewidth, center]{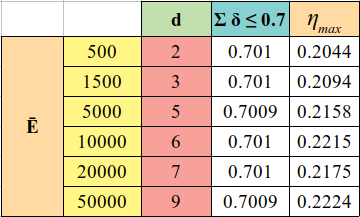}
\end{figure}
Although the \textit{turquoise} column should be less than or equal to $0.7$, at some calculations it turns to be greater due to small numerical noise, stemmed by residues from the GA initialization. Described in details at the following document [\ref{noise}].

\newpage
From here on, we'll rise the number of sections to DoF $ = \{2, 3, 4, 5\}$ : 
\begin{figure}[H]
\centering
\includegraphics[width=0.9 \linewidth, center]{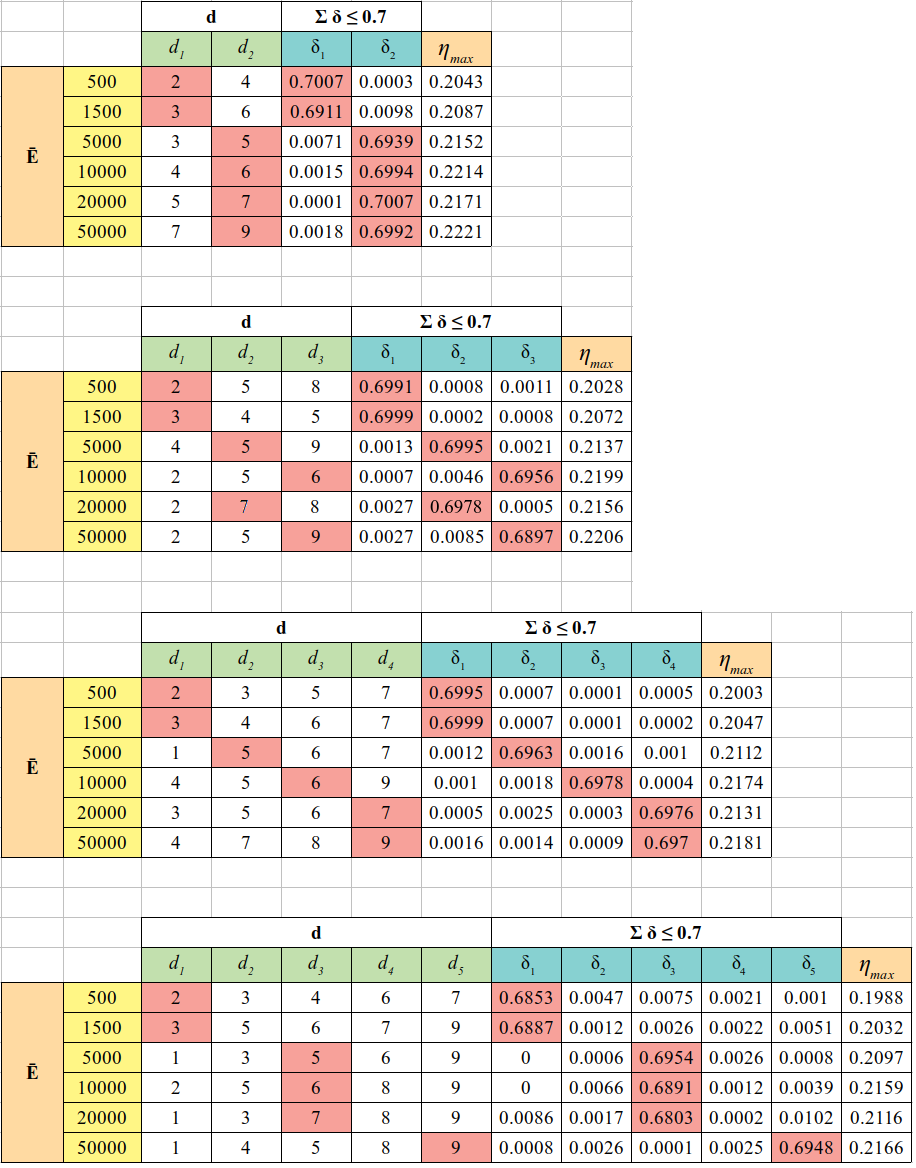}
\end{figure}
The red marked cells are designated to emphasize the repeated pattern of a given section with respect to certain evaporation rate, namely optimal \,$\bar{E}$ .

\newpage
Consider the following executions for DoF $ = \{6, 7, 8, 9\}$ : 
\begin{figure}[H] \label{eta_d_6} 
\centering
\includegraphics[width=1.15 \linewidth, center]{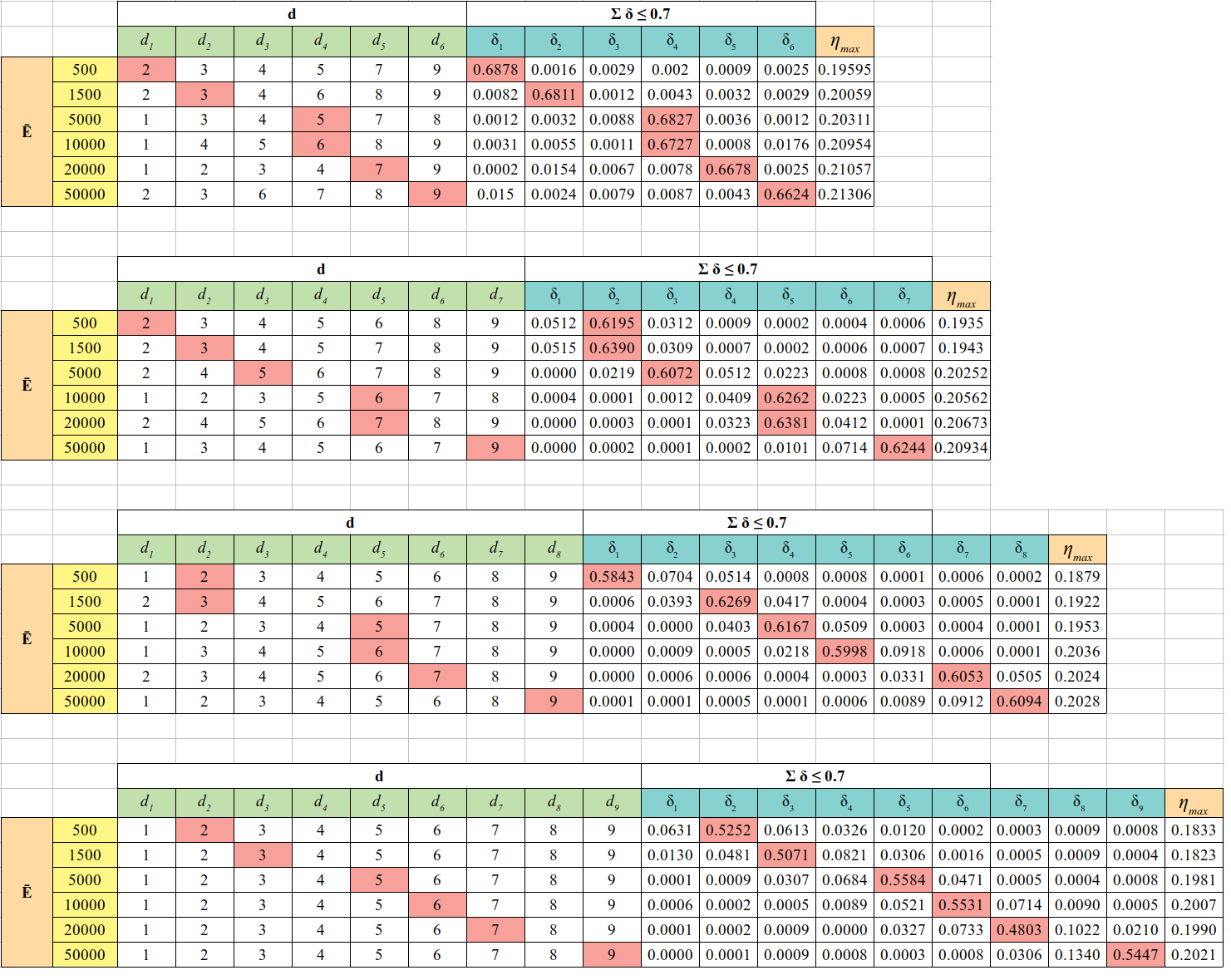}
\end{figure}

\subsubsection{Discussion}
Overall, we can see that across all of the executions, the flame height exhibits values within a bounded range of \,$\eta_{max} \in [ 0.1823 , 0.2224 ]$. Interestingly, the $d^* = \{2, \, 3, \, 5, \, 6, \, 7, \, 9  \}$ pattern appears across all executions, regardless the number of DoF / sections. 

However, the optimization "quality" shows growing signs of decay given more sections to optimize, as more liquid fuel occupies the neighboring sections, on account of a single section (which proved to guarantee optimality).

Sum it in a scatter plot :
\begin{figure}[H]
\centering
\includegraphics[width=1.0 \linewidth, center]{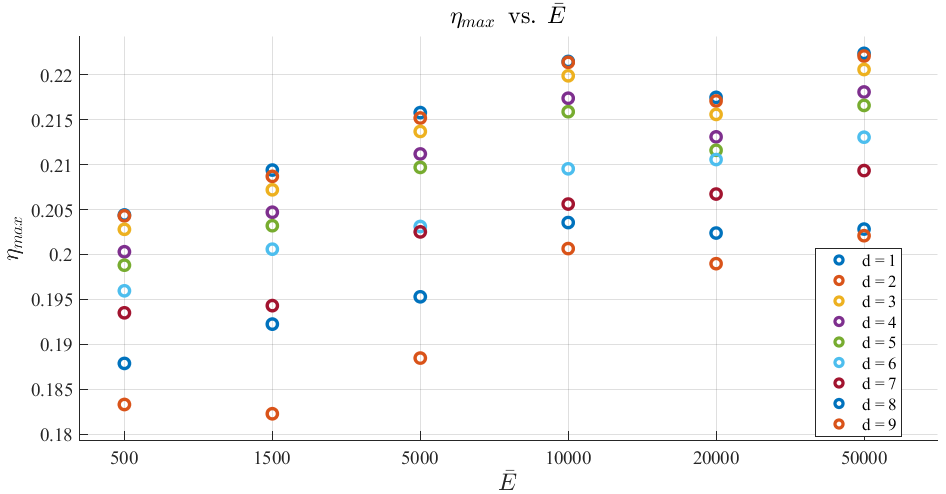}
\end{figure}
Overall, given more sections to optimize we can say that : \\
\quad \, (i) Maximum flame height decreases - \hspace{9mm} $d \ \uparrow \quad \Rightarrow \quad \eta_{max} \hspace{6mm} \downarrow$ \\
\hspace{3.75mm} (ii) Maximum liquid fuel decreases - \hspace{13mm} $d \ \uparrow \quad \Rightarrow \quad \max(\bar{\delta}) \ \downarrow$

\subsection{Flame temperature Optimization}
Similarly to $\eta_{max}$ scheme, we'll do the same with the maximum tip flame temperature :
\begin{figure}[H] 
\centering
\includegraphics[width=0.34 \linewidth, center]{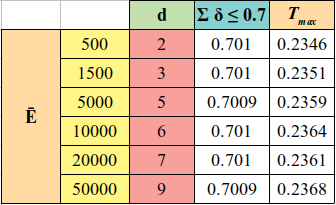}
\end{figure}
Note that the same pattern obtained before, is achieved either here. The reason for that is the close relation of $\eta_{max}$ and $T_{max}$, as mentioned \hyperlink{field_3D}{here}, where $\eta_{max}$ location tends to coincide with $T_{max}$, and vice versa.

\newpage
Rising the number of sections - DoF $= \{2, 3, 4, 5\}$ : 
\begin{figure}[H]
\centering
\includegraphics[width=0.885 \linewidth, center]{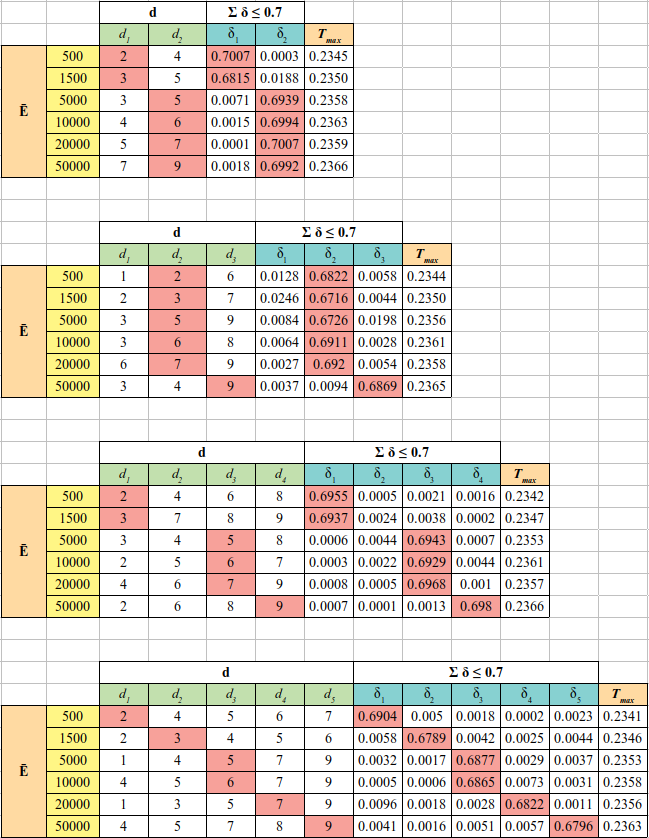}
\end{figure}

\newpage
Consider the following executions for DoF $= \{6, 7, 8, 9\}$ 
\begin{figure}[H] \label{T_d_6} 
\centering
\includegraphics[width=1.15 \linewidth, center]{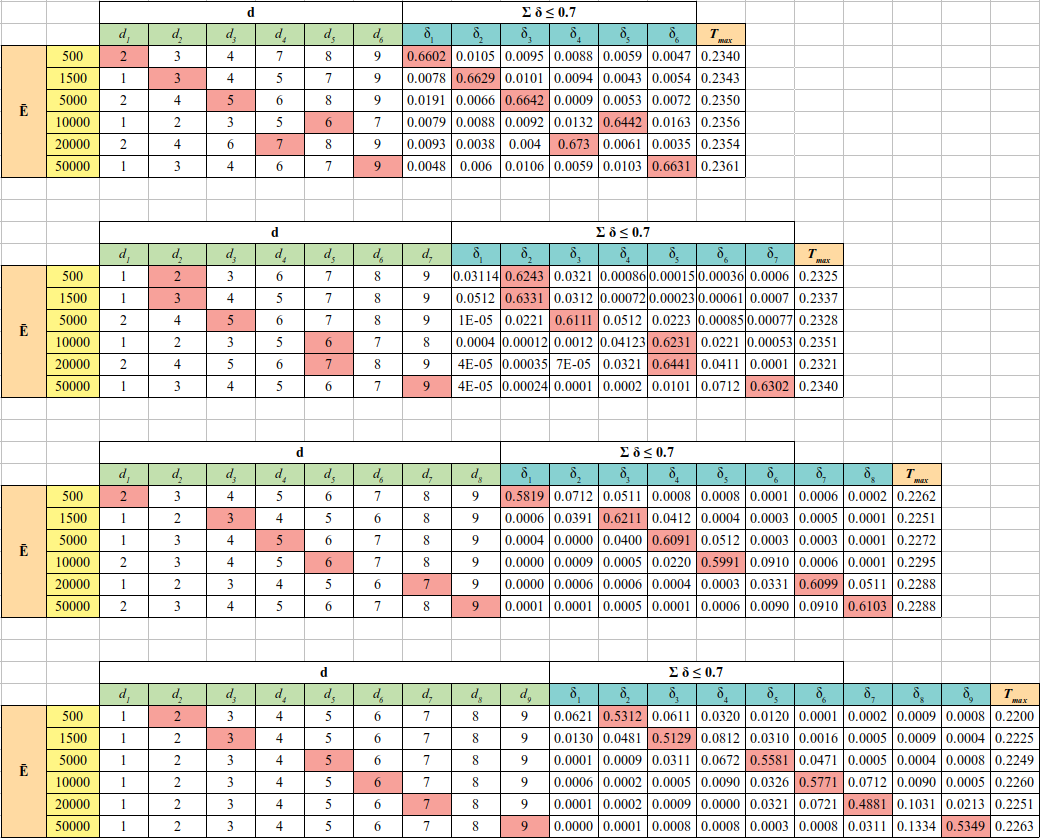}
\end{figure}

Same optimal patterns $d^* = \{2, \, 3, \, 5, \, 6, \, 7, \, 9  \}$ as obtained before, seem to appear here as well. The main difference however, is the $T_{max}$ values whose range is slightly above that of $\eta_{max}$, and the relative differences between the values of $T_{max}$ are smaller than those of $\eta_{max}$ . 

\newpage

\subsubsection{Discussion}
Let us sum the above tables in a scatter plot :

\begin{figure}[H]
\centering
\includegraphics[width=1.0 \linewidth, center]{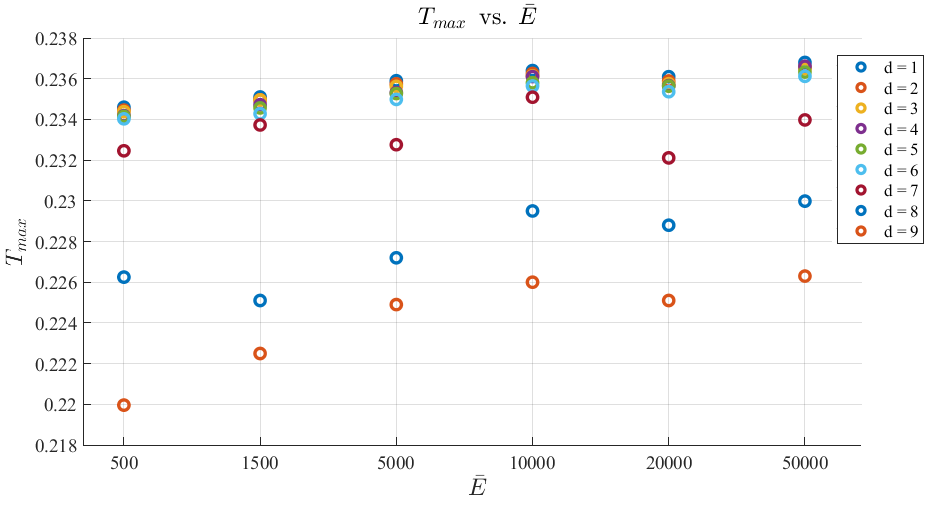}
\end{figure}

Either here the same behavior occurs, when more sections are optimizable : \\
\quad \, (i) Maximum tip flame temperature decreases - \hspace{9mm} $d \ \uparrow \quad \Rightarrow \quad T_{max} \hspace{6mm} \downarrow$ \\
\hspace{3.75mm} (ii) Maximum liquid fuel decreases - \hspace{30.5mm} $d \ \uparrow \quad \Rightarrow \quad \max(\bar{\delta}) \ \downarrow$

However, the range of $T_{max}$ \, is much smaller, and \underline{relative} differences are aggravated as $d \, \uparrow$ .

\subsubsection{Local underperformance}
An interesting phenomenon that can be seen at both \,$\eta_{max}$ \,and \,$T_{max}$ plots, is the lower performances of : $f(\bar{E}_{=2 \cdot 10^4}) < f(\bar{E}_{=10^4})$ . Having no other constraint, shouldn't the reader expect that higher evaporation rates will guarantee better performance ? He definitely should. But one must keep in mind that the \underline{discrete sectioning} of the droplet size into $\max{(d)} \leq 9$, allows a narrow "optimal evaporation zone" for each section size. 

Therefore, the local decrease of $T_{max}(\bar{E}_{=20,000})$ is because of the predetermined $\bar{E}=20,000$ which over-evaporates lower sections (smaller initial droplets size), but on the other hand is too slow for higher sections (larger iDSD), and thus underperforms.

\subsection{Optimization summary}
In this section I executed two optimization scenarios, each composed of a varied DoF, constrained only by \,$\sum_i \delta_i \leq 0.7$ and \,$\bar{E}_{const}$. We saw that the GA was able to find an optimal iDSD, across different parametric configurations. That is to say, that by properly delivering a complex problem to the GA, a near-optimal solution will (eventually) be found.

Moreover, it was \underline{empirically} shown that optimality of both  $\eta_{max}$ and $T_{max}$\, is guaranteed mostly when the liquid fuel occupies only one single section. Namely, optimal performance is guaranteed when the iDSD is monosectional. But why is that ? The answer $\rightarrow$ [\ref{validation}]

The DoF number was found as an important player in the optimization game, as higher DoF decay the results' quality. For that reason, the following scheme will settle for 6 DoF.
 
\subsection{Péclet Number}
In this part I would like to examine the Péclet number influence on a given optimal iDSD, which so far was determined by default as $Pe = 10$ . \\ \vspace{3mm}
\hspace{12mm} \ What is the Péclet Number ?

A class of dimensionless numbers relevant in the study of transport phenomena that denotes the ratio of the advection rate by the diffusion rate driven by an appropriate gradient  [\ref{HT}]. In the context of mass transfer, the Péclet number is the product of the \textit{Reynolds} number and the \textit{Schmidt} number :
\begin{align*}
Pe_R = Re_R \, Sc = \frac{U_{0_g} R}{D_g}
\end{align*}
\hspace{15mm} $\circ$ \ $D_g$\, - Mass diffusion coefficient \\
\hspace{15mm} $\circ$ \ $U_{0_g}$ - Local flow velocity \\
\hspace{15mm} $\circ$ \ $R$ \hspace{1.6mm} - Characteristic length (half external channel)

The next section consist of two scenarios of both \,$\eta_{max}$ and \,$T_{max}$ . \\ Each scenario comprises of two different subsections : \\
(i) \hspace{0.2mm} A general sensitivity test of a\, \underline{monosectional} iDSD to $Pe$ number (for perspective). \\
(ii) A specific sensitivity test of the \underline{polysectional} iDSD obtained at \hyperref[T_d_6]{6 DoF} optimization.

\subsubsection{Pe vs. flame height}

Consider the following $\eta_{max}^{(\sum \delta=0.7)}$ space presenting 3 different monosectionals - $d = \{2, 5, 8 \}$ :
\begin{figure}[H]
\centering
\includegraphics[width=1.1 \linewidth, center]{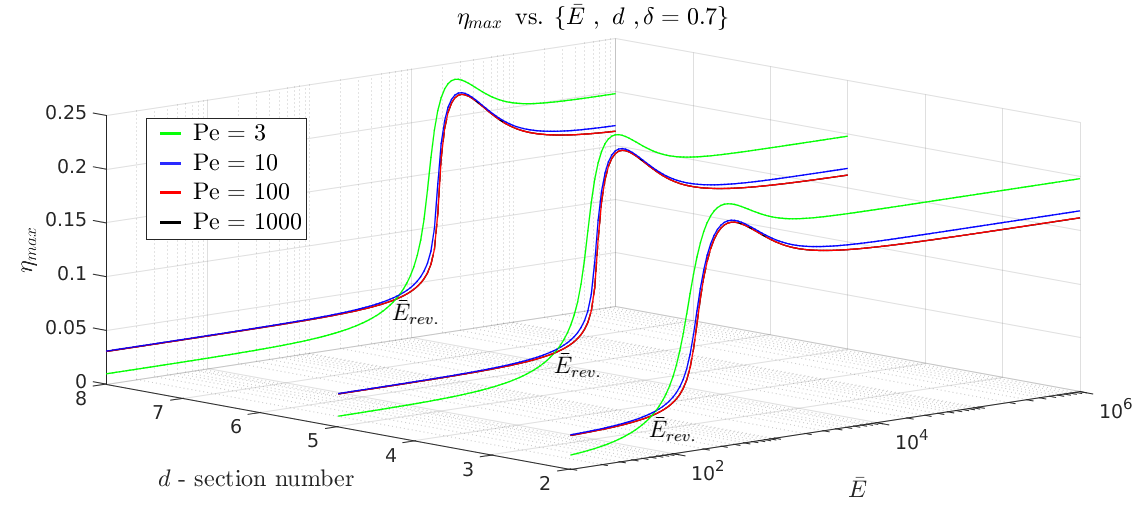}
\end{figure}
Notice the ambivalent relation of $Pe$\, depending on the evaporation rates (rev. $\equiv$ {reversal}) :
\begin{align*}
\begin{cases}
\bar{E} < \bar{E}_{rev.} \ , \ Pe \ \uparrow \quad \Rightarrow \quad \eta_{max} \ \uparrow  \\
\bar{E} > \bar{E}_{rev.} \ , \ Pe \ \uparrow \quad \Rightarrow \quad \eta_{max} \ \downarrow
\end{cases} 
\end{align*}

Now, consider the \underline{polysectional} iDSD obtained at \hyperref[eta_d_6]{$\eta_{max}$} optimization in \underline{6 DoF} :
\begin{figure}[H] \label{eta_Opt}
\centering
\includegraphics[width=1.0 \linewidth, center]{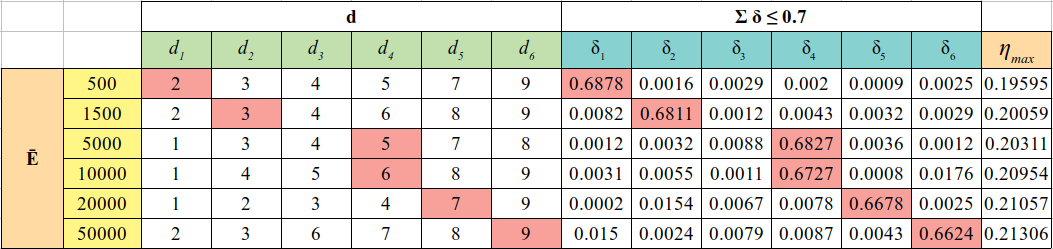}
\end{figure}
Now we shall examine their sensitivity to several $Pe$ numbers, where the rounded rectangle in the bottom left side denotes the gaseous flame \underline{height} sensitivity to $Pe$.

\begin{figure}[H]
\centering
\includegraphics[width=.975 \linewidth, center]{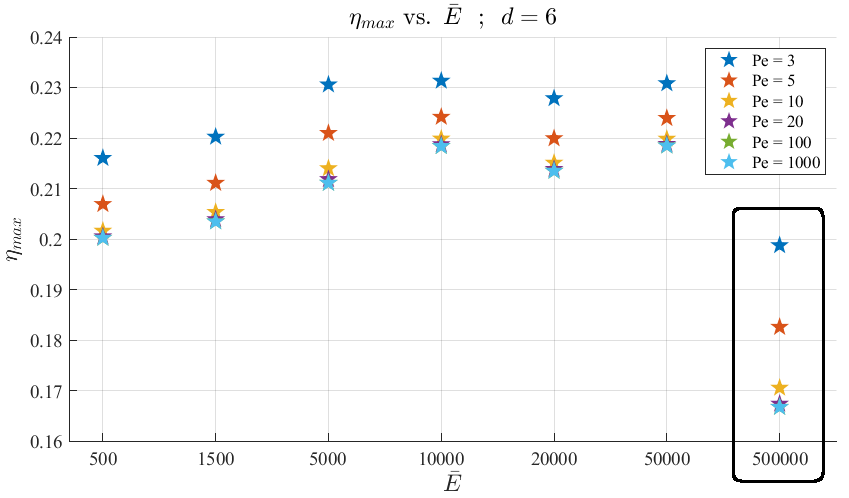}
\end{figure}

\subsubsection{Discussion}
In the first figure we can see a general solution space where some $\bar{E}$ regions benefit with some $Pe$ while others are harmed. Small $Pe$ numbers at low evaporation rates ( $\bar{E}_{< rev.}$ ) cause an \underline{attenuation} reaction of $\eta_{max}$ . But after passing $\bar{E}$ reversal point, small $Pe$ numbers \underline{enhance} \,$\eta_{max}$. The phenomenon is even more emphasized towards higher evaporation rates $\bar{E} \rightarrow \infty$ (gaseous flame height), as marked in the rounded rectangle.

Overall, it makes sense, as the reversal point denotes the $\bar{E}$\, threshold from which the combustion becomes efficient and start rising. Additionally, the difference between $Pe = 100$ and $Pe = 1000$ seems almost negligible, for any evaporation rate. In the second (above) figure, the ambivalent policy no longer exists, and instead we get :
\begin{align*}
Pe \ \downarrow \quad \Rightarrow \quad \eta_{max} \ \uparrow \ \quad \forall \quad \bar{E}
\end{align*}
The reader may wonder how is that possible ? \\ \label{opt_question}
The answer goes back to the \underline{optimal} iDSD, as stated \hyperref[eta_Opt]{here}. Although being optimized as a polysectional, about \,$ \approx 96 \, \%$ of the liquid fuel is concentrated in \textbf{one section}, making it as if it was a monosectional iDSD. Knowing that, we can see that all these optimal iDSDs take place \underline{after} the reversal points, where smaller $Pe$ numbers \underline{benefits} with \,$\eta_{max}$ . 

\subsubsection{Pe vs. flame temperature}

Consider the following $T_{max}^{(\sum  \delta=0.7)}$ space presenting 3 different monosectionals - $d = \{2, 5, 8 \}$ :
\begin{figure}[H]
\centering
\includegraphics[width=1.05 \linewidth, center]{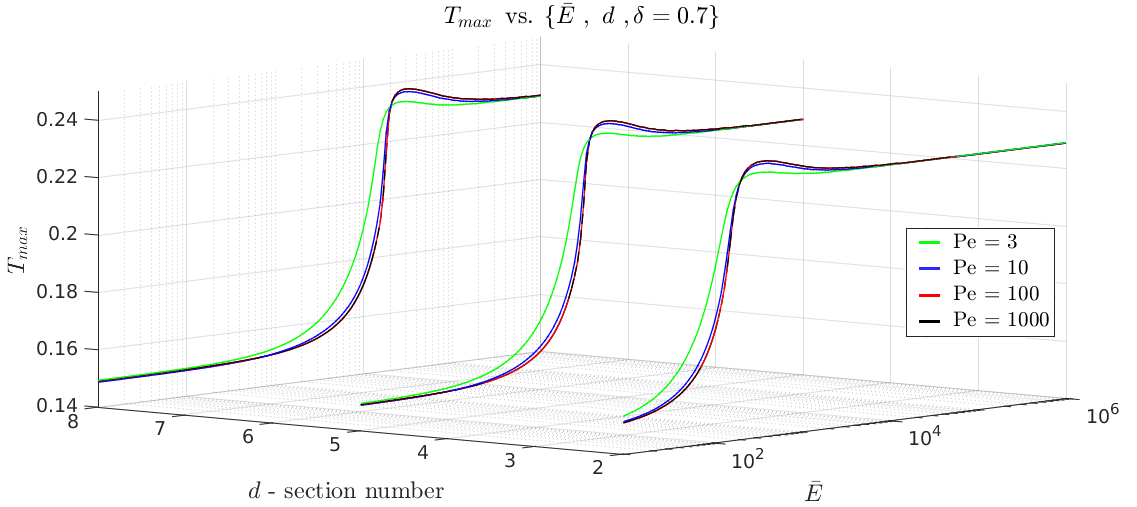}
\end{figure}

The ambivalent relation seen here seems to be completely \textbf{opposite} to the previous one :
\begin{align*}
\begin{cases}
\bar{E} < \bar{E}_{rev.} \ , \ Pe \ \uparrow \quad \Rightarrow \quad T_{max} \ \downarrow \\
\bar{E} > \bar{E}_{rev.} \ , \ Pe \ \uparrow \quad \Rightarrow \quad T_{max} \ \uparrow
\end{cases} 
\end{align*}
Looking closer we can see that the reversal point ( $\bar{E}_{rev.}$ ) is closer to the global maximum :
\begin{figure}[H]
\centering
\includegraphics[width=1.035 \linewidth, center]{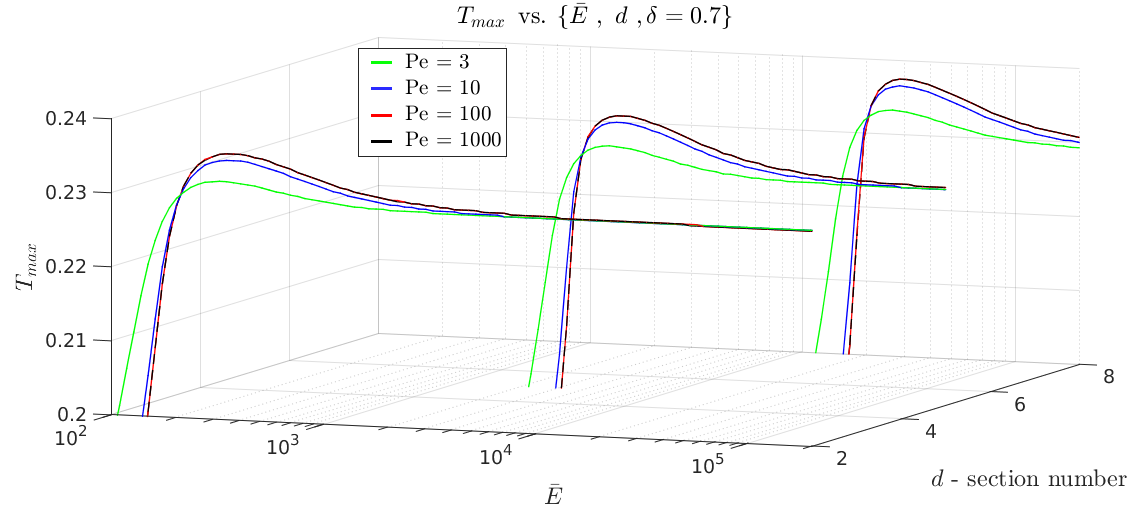}
\end{figure}

Consider the \underline{polysectional} iDSD obtained at \hyperref[T_d_6]{$T_{max}$} optimization at 6 DoF :
\begin{figure}[H]
\centering
\includegraphics[width=1.1 \linewidth, center]{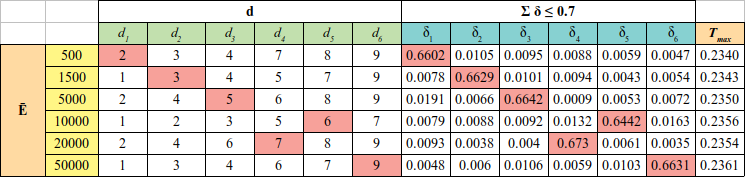}
\end{figure}
Now we shall examine this iDSD sensitivity to several $Pe$ numbers, where the rounded rectangle in the bottom left side denotes the gaseous flame \underline{temperature} sensitivity to $Pe$.
\begin{figure}[H]
\centering
\includegraphics[width=1.0 \linewidth, center]{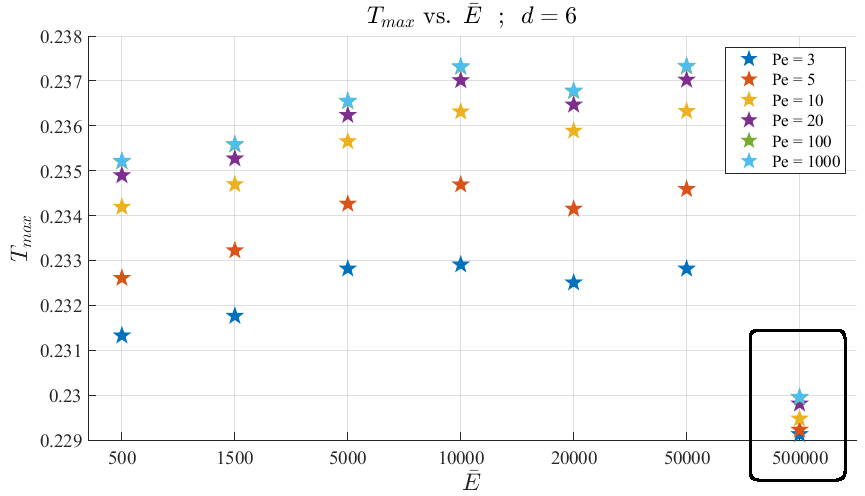}
\end{figure}

As mentioned above the overall policy of $T_{max}$ is completely \textbf{opposite} to that of $\eta_{max}$ . However, the relative differences are less dramatic :
\begin{align*}
\frac{\eta_{max}^{(Pe=3)}}{\eta_{max}^{(Pe=1000)}} \approx 1.1 \quad > \quad \frac{T_{max}^{(Pe=1000)}}{T_{max}^{(Pe=3)}} \approx 1.015
\end{align*}

Unlike the gaseous $\eta_{max}^{(E \rightarrow \infty)}$, here the gaseous $T_{max}^{(E \rightarrow \infty)}$ shows a weak response to $Pe$ .

\subsubsection{Discussion}
The first figure exhibited a general solution space of $T_{max}$ where some $\bar{E}$ regions benefited with $Pe$ while others were harmed. In the $\bar{E}_{< rev.}$ region, small $Pe$ numbers \underline{enhanced} $T_{max}$, but after passing the reversal point ($\bar{E}_{rev.}$), small $Pe$ numbers seemed to \underline{attenuate} \,$T_{max}$. Contrarily to $\eta_{max}$ , the gaseous flame temperature obtained at ( $\bar{E}_{\rightarrow \infty} \gg \bar{E}_{rev.}$ ) showed rather small relative differences between different $Pe$ numbers. In conclusion :
\begin{align*}
\begin{cases}
\bar{E} > \bar{E}_{rev.} \ , \ Pe \ \uparrow \quad \Rightarrow \quad \eta_{max} \ \downarrow \quad T_{max} \ \uparrow \\
\bar{E} > \bar{E}_{rev.} \ , \ Pe \ \downarrow \quad \Rightarrow \quad \eta_{max} \ \uparrow \quad T_{max} \ \downarrow
\end{cases} 
\end{align*}

Also here I will make use of the same explanation as brought \hyperref[opt_question]{before}, regarding the optimal combinations satisfying $\bar{E} > \bar{E}_{rev.}$ , where higher $Pe$ number \underline{benefits} with \,$T_{max}$ . 

As an auxiliary argument, consider the following reference from the thesis (p. 117) [\ref{ga33}], presenting the extinction maps that describe the flame's sensitivity to the iDSD and $Pe$ : \vspace{4mm}

\begin{figure}[H]
\centering
\includegraphics[width=1.2 \linewidth, center]{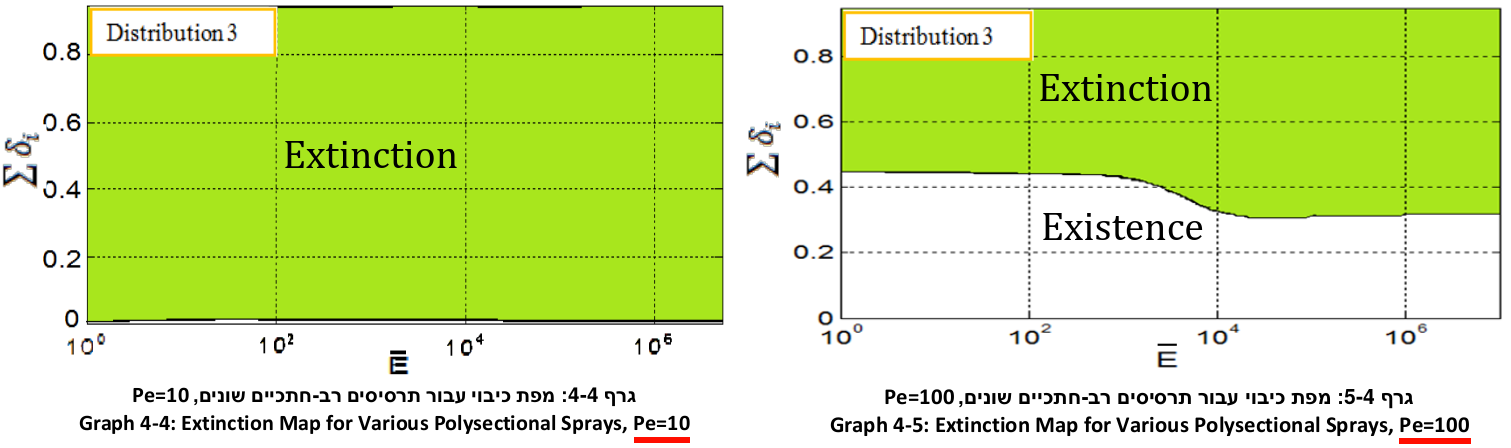}
\end{figure}

Without deep diving into Liñán's diffusion flame theory [\ref{linan}], and without loss of generality, we can \underline{carefully} say that higher $Pe$ numbers reduce the extinction regions. 

Equivalently, that is to say that higher $Pe$ numbers expand the existence regions of the flame itself. Note that this example's iDSD is arbitrary but it still provides a useful qualitative information that supports my findings. 

\subsection{Empirical validation} \label{validation}

The above executions have shown the monosectional superiority in achieving optimal performances. In attempt to explain that, I would like to provide another point of view for analysis, that might shed some light on that behavior and utilize as a reliable sanity check.

Consider the following image, presenting seven different initial distributions containing the same total liquid fraction. The first iDSD is monosectional, but as we go forth the standard deviation increases and spreads to more sections : \newline

\begin{figure}[H]
\centering
\includegraphics[width=0.75  \linewidth, center]{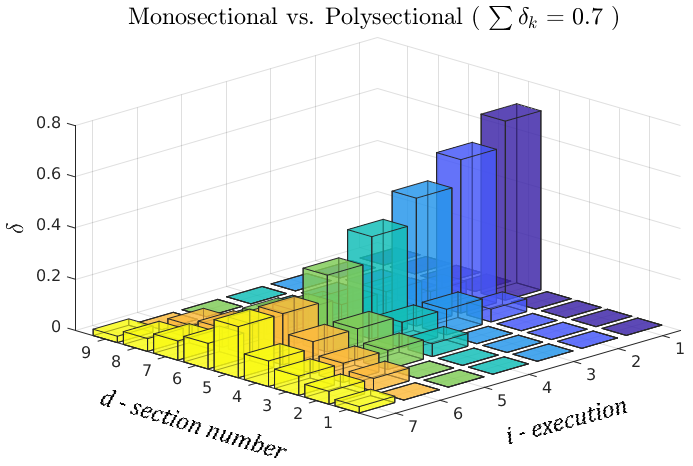}
\end{figure}
The 1st iDSD ($:=$'data1') contains only one section \,$\delta_{d=5}=0.7$ . But 2nd iDSD ('data2') :
\begin{align*}
\delta_{d=4} + \delta_{d=5} + \delta_{d=6} = 0.05 + 0.6+0.05 = 0.7
\end{align*}
And so on until reaching the seventh iDSD ('data7') that consists of \underline{all nine} sections in a normal-like distribution, namely all sections are fueled.

\newpage

Using these for plotting the maximum flame height as a function of the evaporation rate :
\begin{figure}[H]
\centering
\includegraphics[width=0.85  \linewidth, center]{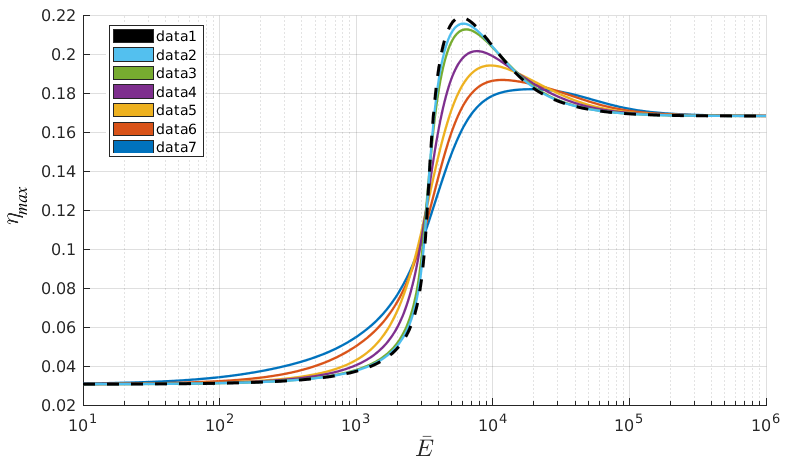}
\end{figure}
We can see that the highest flame is obtained for monosectional (dashed \textit{black}), and the more the distribution is spread (more sections are "occupied"), so \,$\eta_{max}$ turns lower. 

However, one may come up and claim that at a certain closed interval, the monosectional iDSD actually performs the poorest, while others are optimal :
\begin{figure}[H]
\centering
\includegraphics[width=0.63  \linewidth, center]{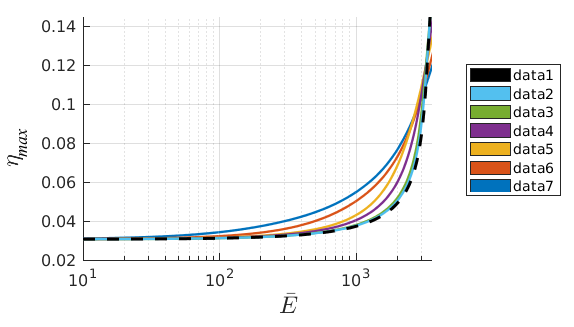}
\end{figure}
Indeed, the polysectional iDSDs do perform better on lower $\bar{E}$, as the distribution spread wider. However, that argument is only half true. Let us add two more monosectional distributions to the current figure. 

The dashed line from left is a monosectional of $\delta_{d=2}$, and the right one is of $\delta_{d=7}$ :

\begin{figure}[H]
\centering
\includegraphics[width=0.815  \linewidth, center]{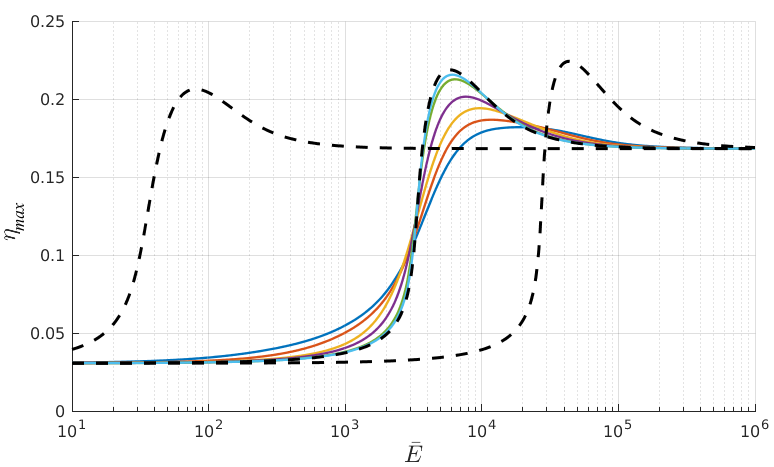}
\end{figure}
Note that any non-monosectional distribution is situated beneath some of the monosectional. Adding more monosectionals and we get the full picture :
\begin{figure}[H]
\centering
\includegraphics[width=0.875  \linewidth, center]{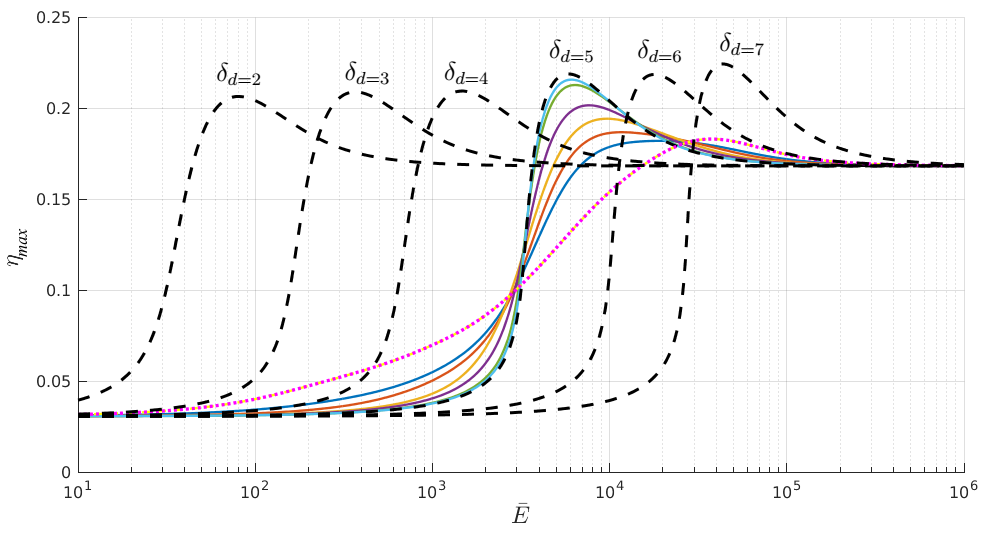}
\end{figure}
The conclusion is clear, any polysectional iDSD is always bounded between neighboring monosectionals. Therefore the latter performs better at any evaporation rate. As a touchstone, I added a \underline{uniform} distribution upon all sections  - $\delta_{d=[1:9]} = \frac{0.7}{9}$ \ ( dotted \textit{pink} ) . 

\newpage

\section{Conclusions}

In this research project I integrated a combustion model within a Genetic Algorithm in an innovative manner that was not performed before. The main research question : \\
\hspace{20mm} "What form of an iDSD will guarantee the optimal flame performances ?"

Was reviewed from different point of views, at first to gain a bird's-eye view and afterwards by a closer inspection. The optimization scheme left no room for doubt about the optimal distribution for both \,$\eta_{max}$  or  $T_{max}$ , which was found to be the monosectional iDSD. 

Moreover, a sensitivity test of the model with respect to Péclet number was provided, followed by an empirical validation that supported the research findings.

To conclude, the GA was successfully harnessed for the sake of the developed engineering problem, thereby reinforcing the idea of problems that were so far non-optimizable, can now be optimized and provide an heuristic optimal solution.

\subsection{Further Work}

Along my working process I happened to think through several directions for future work :

\quad $\circ$ \ \underline{Experimentally} - Given appropriate equipment, do the optimal values found converge  \\ \hspace{8mm} with real laboratory experiments ?

\quad $\circ$ \ \underline{Technically} - Do the number of sections (9 by default) have any influence upon the \\ \hspace{8mm} results ? Would more sections necessarily lead to a more continuous results space ?

\quad $\circ$ \ \underline{Complexity Analysis} - Long and costly computation process can be analyzed, in order  \\ \hspace{8mm} to point to bottlenecks and to make the algorithm more efficient.

\quad $\circ$ \ \underline{More approaches} - Nowadays, the GA main competitor is the AI's top notch approach,  \\ \hspace{8mm} the Reinforcement Learning. In this approach, an agent in a given environment (input  \\ \hspace{8mm} setup), reflects the user interest to optimize a certain value function. The agent  \\ \hspace{8mm} improves his actions due to external rewards with respect to his actions. Over time he \\ \hspace{8mm} attempts to arrive at an optimal state, namely a set of actions that reward him most, \\ \hspace{8mm} thereby guaranteeing optimality.

\newpage


\newpage

\section*{Appendices} \addcontentsline{toc}{section}{Appendices}

\subsection*{\hypertarget{app_extremum}{Appendix A} - Extremum definition} \addcontentsline{toc}{subsection}{Appendix A}

Consider the following \underline{2D} $^{[\hyperlink{opt_A1}{A_{\text{1}}}]}$ differentiable function $f(x, y)$ , that satisfies $f : X \rightarrow \mathbb{R}$ \,:
\begin{figure}[H]
\centering
\includegraphics[width=0.975 \linewidth, center]{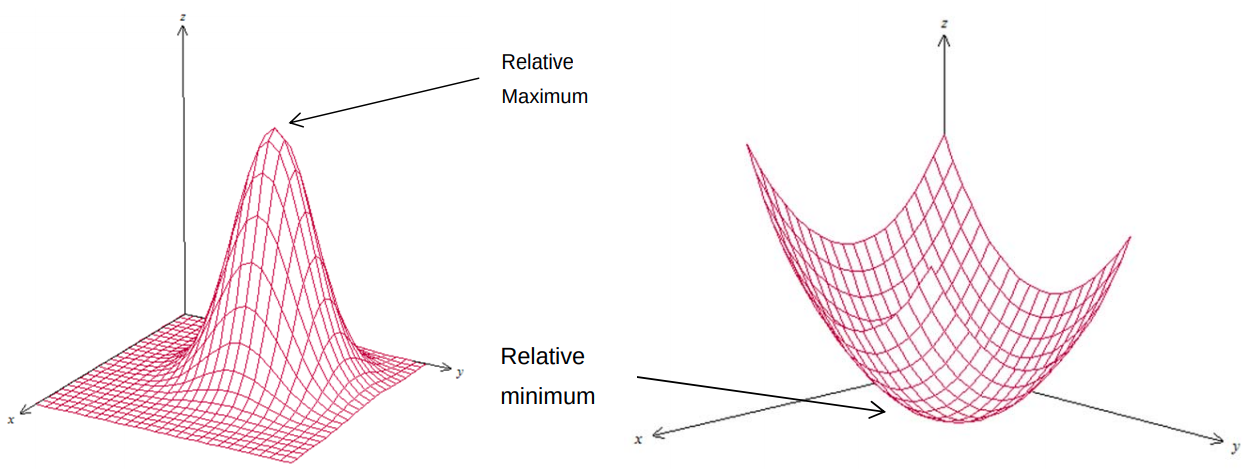}
\end{figure}
Any point within the domain ($X \in \mathbb{R}^2$) can be calculated by a desired function, and then projected onto a new space \,$\big( x, y, f(x, y) \big)  \in \mathbb{R}^{2+1}$. The global maximum of such function $f$\, is the point $\big( c, d, f(c, d) \big)$, if there exists some region surrounding $(c, d)$ for which $^{[\hyperlink{opt_A2}{A_{\text{2}}}]}$ :
\begin{align*}
f(x, y) \ \leq \ f(c, d) \quad \forall \quad (x, y) 
\end{align*}
A function of two variables \,$f$\, has a critical point at $\bar{x}_0 = (c, d)$ if :
\begin{align*}
f(c, d)_{x} = 0 \quad \text{and} \quad f(c, d)_{y} = 0 
\end{align*}
Such that \underline{any} critical point ( $\bar{x}_0 \in \bar{\textbf{x}}_0 $ ) within the domain will satisfy :
\begin{align*}
\nabla \, f( \bar{\textbf{x}}_0 ) \, = \, 0
\end{align*}
The \underline{global extremum} is said to be point - \,$\bar{x}_0^*$ \,, which satisfies one of the following :
\begin{align*} \label{global_ext}
\min_x f(X) \quad : \quad f(\bar{x}_0^*) \leq f(\bar{\textbf{x}}_0) \\
\max_x f(X) \quad : \quad f(\bar{x}_0^*) \geq f(\bar{\textbf{x}}_0) \\
\end{align*}

\subsection*{n-dimensional example}
Given a more general case of a several real variables function that associates an arbitrary n-dimensional point of \, $\bar{{x}} \in \mathbb{R}^n$ \, to \, $f : X \rightarrow \mathbb{R}$ . For illustration, a 3D function $^{[\hyperlink{opt_A3}{A_{\text{3}}}]}$ :
\begin{figure}[H]
\centering
\includegraphics[width=0.425 \linewidth, center]{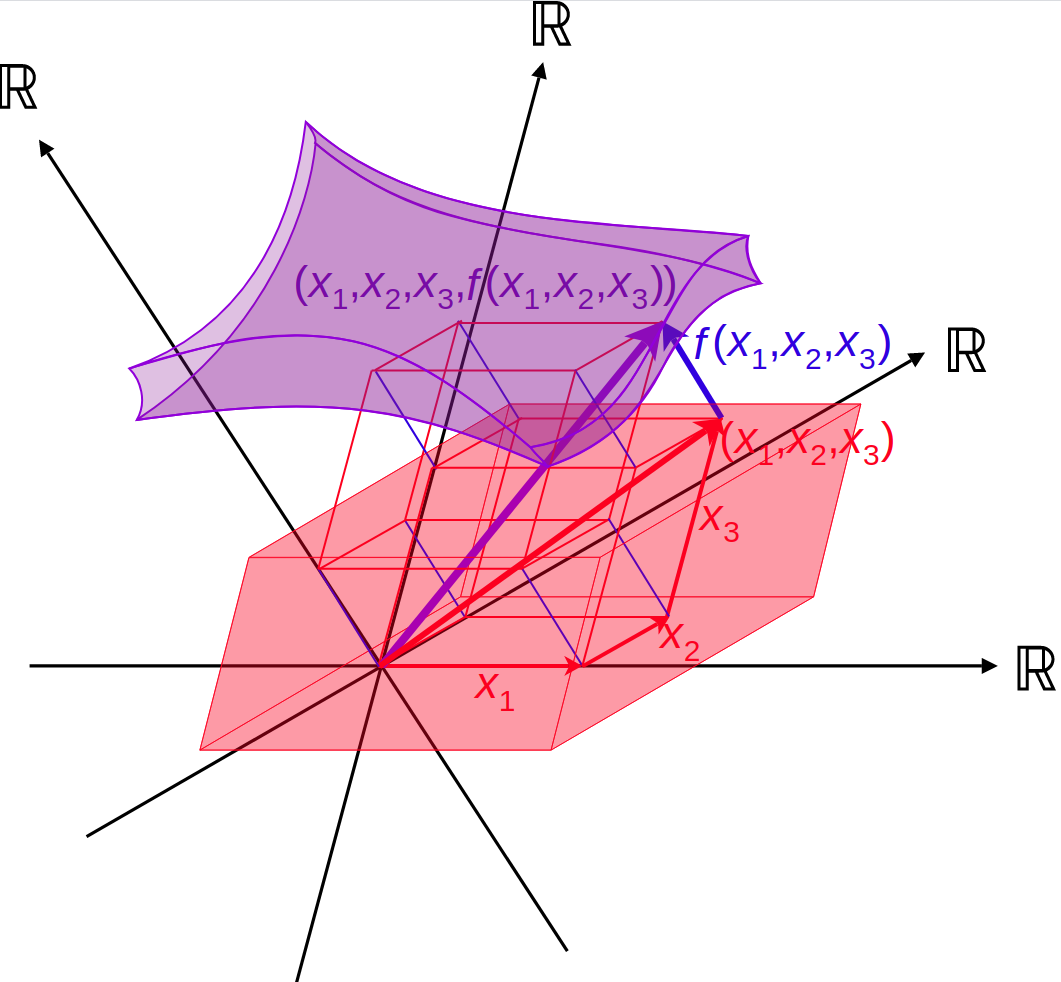}
\caption*{\textit{red} - domain \tab \textit{purple} - image}
\end{figure}
These points can be calculated by an appropriate function such as $f (\, \overbrace{x_1, \hdots, x_n}^{\bar{x}} \,)$, and then projected onto $\big( \bar{{x}}, f(\bar{\textbf{x}}_0 \big) \in \mathbb{R}^{n+1}$ space. Derivation of $f$ provides a system of $n$ equations :
\begin{align*}
\nabla \, f = \Big( \frac{\partial f}{\partial x_1} \, , \ \hdots \ , \,  \frac{\partial f}{\partial x_n} \Big) = \bar{0}
\end{align*}
Whereas either here, \,$\bar{\textbf{x}}_0$\, is the set of \underline{critical} points that nullify the gradient :
\begin{align*}
\nabla \, f( \bar{\textbf{x}}_0 ) \, = \, 0
\end{align*}
Similarly to before, the global extrema would yield :
\begin{align*}
\min_x : \ f(\bar{x}_0^*) \leq f(\bar{\textbf{x}}_0)  \quad \Leftrightarrow \quad \max_x : \  f(\bar{x}_0^*) \geq f(\bar{\textbf{x}}_0)
\end{align*}

\textbf{References}

$\bullet$ \ \hypertarget{opt_A1}{Boris P.}, Baranenkov, G. Moscow University (1964). Problems in mathematical analysis.  \\
$\bullet$ \ \hypertarget{opt_A2}{Stewart, James} (2008). \href{https://archive.org/details/calculusearlytra00stew_1}{Calculus: Early Transcendentals (6th ed.)}. Brooks/Cole. \\
$\bullet$ \ \hypertarget{opt_A3}{Craig A.} Tovery, Georgia Institute of Technology (2010). \href{https://www.extremeoptimization.com/Documentation/Mathematics/Optimization/Multidimensional-Optimization.aspx}{Multidimensional Optimization}

\subsection*{\hypertarget{app_convex}{Appendix B} - Convexity vs. concavity}   \addcontentsline{toc}{subsection}{Appendix B}

\textbf{Definition B.1} - A real-valued function $f$ is considered \textbf{convex} if the line segment between any two points on graph of the function $f(x)$ lies above or on the graph. 

\textbf{Definition B.2} - A subset $\mathcal{C}$ is considered \textbf{convex set} if, with any two points, it contains the \underline{whole} line segment that joins them.

Let $X$ be a convex set in a vector space and let $f:X \rightarrow \mathbb{R}$ be a function. $f$ is convex if :
\begin{align*}
\forall \ x_1, x_2 \ \in \ X \ \forall \ t \in [0, 1] : \tab 
f( t x_1 + (1-t)x_2) \leq tf(x_1) + (1-t)f(x_2)
\end{align*}
\textbf{\hypertarget{convex_concave}{Definition B.3}} - $f$ is said to be \textbf{concave} if \,($-f$) is convex (negative of a convex function).

Knowing the function's type dictates the objective function's type (minimize / maximize).

\subsubsection*{Optimization type}

An optimization problem is said to be convex if its objective function is a convex function$^{[\textbf{2.1}]}$ and the constraint set is a convex set$^{[\textbf{2.2}]}$. The optimization task is to find the global extremum (convex = minimum, concave = maximum), and is expressed typically as :
\begin{align*}
&\underset{\mathbf{x}}{\operatorname{minimize}}& & f(\mathbf{x}) \\
&\operatorname{s.t} \ x \in \mathcal{C}
& &g_i(\mathbf{x}) \leq 0, \quad i = 1, \dots, m \\
&&&h_i(\mathbf{x}) = 0, \quad i = 1, \dots, p,
\end{align*}
Where $f$ and $\mathcal{C} := \{ g_1, \hdots, g_m, \, f_1, \hdots, f_p \}$ are convex. \\
The important characteristic we get is that the \underline{local} minimum is a \underline{global} minimum.

\textbf{Definition B.4} - $\mathcal{C}$ is said to be a \textbf{non-convex set} in the presence of at least one single \underline{uncontained} convex combination. Analogously, \textbf{non-concave set} are with uncontained concave combination. Conventionally, \underline{both} terms are referred as non-convex.

An optimization problem that violates either one of these conditions $^{[\textbf{B.1}], [\textbf{B.2}]}$, i.e. utilizes a non-convex objective function, or a non-convex constraint set $^{[\textbf{B.4}]}$, is considered as a \textbf{non-convex} optimization problem.

\begin{figure}[H]
\centering
\includegraphics[width=1.0 \linewidth, center]{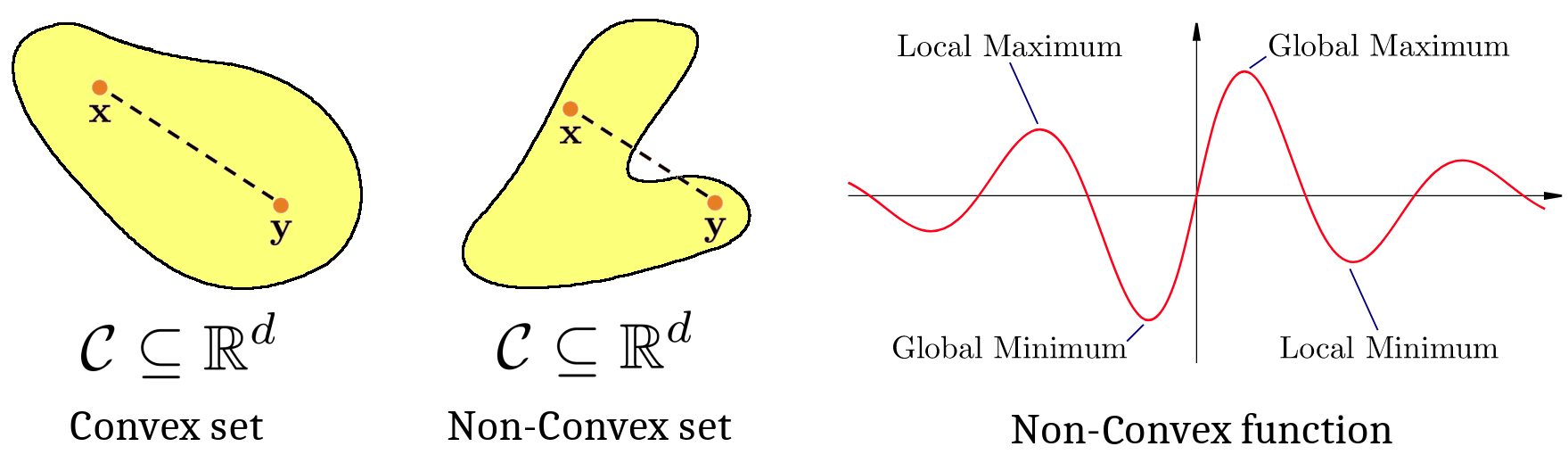}
\end{figure}

A non-convex optimization may have multiple \underline{locally} optimal points such that identifying whether the problem has no solution or if the solution is global, is a challenge per se. 

\vspace{1cm}

\textbf{References}

$\bullet$ \ Meyer, Robert (1970), The validity of a family of optimization methods [\href{https://minds.wisconsin.edu/bitstream/handle/1793/57508/TR28.pdf?sequence=1}{link}] \\
$\bullet$ \ Zălinescu, C. (2002). Convex analysis in general vector spaces [\href{https://archive.org/details/convexanalysisge00zali_934/page/n11/mode/2up}{link}] \\
$\bullet$ \ Kjeldsen, Tinne Hoff. (2006), Convexity and Mathematical Programming [\href{https://web.archive.org/web/20170811100026/http://www.mathunion.org/ICM/ICM2010.4/Main/icm2010.4.3233.3257.pdf}{link}] \\
$\bullet$ \ Shashi Kant Mishra (2011), Topics and applications in Nonconvex Optimization [\href{https://books.google.co.il/books?id=wABn1WoOFR4C&pg=PA89&lpg=PA89&dq=c+is+said+to+be+non-convex+set&source=bl&ots=mZuYjHJ0ws&sig=ACfU3U2qxHRGOgd7kW1M7QGqSuP8buOtfQ&hl=
en&sa=X&ved=2ahUKEwjjxqOvv43qAhUGBGMBHWYzCgIQ6AEwCnoECAcQAQ#v=onepage&q=c%20is%20said%20to%20be%20non-convex%20set&f=false}{link}]

\subsection*{\hypertarget{code_access}{Appendix C} - Code Access} \addcontentsline{toc}{subsection}{Appendix C}

As part of my worldview, not only \underline{ideas} should be accessible to everyone who desires, but also their \underline{implementation} means. Optimally, alongside a clear installation instructions and an execution scheme. The world today is flat, very much thanks to the \href{https://en.wikipedia.org/wiki/Open-source_model}{open source} culture at both academy and industry. 

Aside from global access, it promotes a transparency climates where the laymen is capable of challenging the biggest researchers by validating their results, and check whether it corresponds to their publications. Using the most popular platform, I would like to refer the reader to \,\textit{GitHub.com}\, website where the project's contents can be found :
\begin{center}
\url{https://github.com/Daniboy370/Masters-Project} \\
\vspace{0.5cm}
- $fin$ -
\end{center}

%

\end{flushleft}

\begin{thebibliography}{9} \addcontentsline{toc}{section}{References}

\bibitem{c6} \label{c6}
Beyer, H. G. (2001). The theory of evolution strategies. \href{https://www.springer.com/gp/book/9783540672975}{Springer Science}


\bibitem{c_optima} \label{c_optima}
Paul mcgregor (2006). Relative Minimums and Maximums, \href{https://tutorial.math.lamar.edu/Classes/CalcIII/RelativeExtrema.aspx}{Calculus III course}, Lamar University, Texas


\bibitem{c7} \label{c7}
Darrell Whitley (1994). A genetic algorithm tutorial. Computer Science Department, Colorado State University. \href{https://www.cs.colostate.edu/~genitor/MiscPubs/tutorial.pdf}{Fort Collins, CO 80523, USA}


\bibitem{ga0} \label{ga0}
Runhe Huang (1995). Evolving Prototype Rules and Genetic algorithm
in a Combustion Control. \href{https://ieeexplore.ieee.org/document/465834?section=abstract}{1995 IEEE-IAS}, International Conference on Industrial Automation and Control Conference.


\bibitem{ga1} \label{ga1}
B. Danielson J. $\&$ Foster D. Frincke (1998). Using Genetic Algorithms to Breed a Combustion Engine. \href{https://ieeexplore.ieee.org/document/699722}{IEEE World Congress} on Computational Intelligence, 1998, Anchorage, Alaska, USA


\bibitem{ga2} \label{ga2}
Wolfgana Polifke, Weiqun Geng $\&$ Klaus Dobbeling (1998). Optimization of Rate Coefficients for Simplified Reaction
Mechanisms with Genetic Algorithms. \href{https://www.sciencedirect.com/science/article/abs/pii/S0010218097002125}{Combustion and Flame}, 113(1/2), 119–134.


\bibitem{f11} \label{f11} 
S.D. Harris, Elliott, L., Ingham, D. B., M. Pourkashanian $\&$ C. W. Wilson, (2000). The optimisation of reaction rate parameters for chemical kinetic modelling using genetic algorithms. In ASME Turbo Expo 2002: \href{https://asmedigitalcollection.asme.org/GT/proceedings-abstract/GT2002/36061/563/295726}{Power for Land, Sea, and Air} (pp. 563-572).


\bibitem{f12} \label{f12}
G. R. Vossoughi $\&$ Siavash Rezazadeh (2005). Optimization of the Calibration for an Internal Combustion Engine Management System Using Multi-Objective Genetic Algorithms. Evolutionary Computation, 2005. \href{https://www.researchgate.net/publication/4201526_Optimization_of_the_calibration_for_an_internal_combustion_engine_management_system_using_multi-objective_genetic_algorithms}{The 2005 IEEE Congress} on, Volume: 2


\bibitem{ga3} \label{ga3}
C. D. Rose, S. R. Marsland $\&$ D. Law (2009). Optimisation of the Gas-Exchange System of Combustion Engines by Genetic Algorithm. \href{https://ieeexplore.ieee.org/document/4804021}{2009 4th International Conference} on Autonomous Robots and Agents.


\bibitem{ga33} \label{ga33}
Shtauber, I. \& Greenberg, J.B. (2010), A study of Polydisperse Spray Diffusion Flames and their Extinction in Co-flow. \href{http://www.graduate.technion.ac.il/Theses/Abstracts.asp?Id=24930}{Final Paper towards M.Sc in Aerospace Engineering}


\bibitem{f1} \label{f1}
Nejra Sikalo, Olaf Hasemann, Christof Schulz, Andreas Kempf $\&$ Irenaus Wlokas (2015). A Genetic Algorithm-Based Method for the Optimization of Reduced Kinetics Mechanisms. \href{https://www.researchgate.net/publication/282270904_A_Genetic_Algorithm-Based_Method_for_the_Optimization_of_Reduced_Kinetics_Mechanisms}{International Journal of Chemical Kinetics 47}


\bibitem{f2} \label{f2}
Carolyn R. Kaplan, Alp Ozgen $\&$ Elaine S. Oran (2017).Chemical-diffusive models for flame acceleration and
transition-to-detonation: genetic algorithm and optimisation procedure. \href{https://arxiv.org/abs/1709.00096}{Combustion Theory and Modelling, 2019}


\bibitem{ga4} \label{ga4}
Hongguang Pan, Weimin Zhong, Zaiying Wanga $\&$ Guoxin Wanga (2017). Optimization of industrial boiler combustion control system based on genetic algorithm. \href{https://www.sciencedirect.com/science/article/pii/S0045790617325302?via%3Dihub}{Computers and Electrical Engineering} 70 (2018) 987–997


\bibitem{ga5} \label{ga5}
Jie Liua,b, Biao Maa $\&$ Hongbo Zhaoa (2019). Combustion parameters optimization of a diesel/natural gas dual fuel engine using genetic algorithm. \href{https://www.sciencedirect.com/science/article/pii/S0016236119317193?via%3Dihub}{Fuel 260 (2020) 116365}


\bibitem{ga7} \label{ga7}
Yiding Zhao, Qinghe Wu, Heng Li, Shuhua Ma $\&$ Ping He (2019). Optimization of Thermal Efficiency and Unburned Carbon in Fly Ash of Coal-Fired Utility Boiler via Grey Wolf Optimizer Algorithm. \href{https://ieeexplore.ieee.org/document/4804021}{2010 International Conference} on Electrical and Control Engineering

\bibitem{ga8} \label{ga8}
Burke, S. P., and T. E. W. Schumann. "Diffusion flames." \href{https://pubs.acs.org/doi/pdf/10.1021/ie50226a005}{Industrial \& Engineering Chemistry} 20.10 (1928): 998-1004.


\bibitem{ga9} \label{ga9}
Greenberg, J.B., "The Burke-Schumann Diffusion Flame Revisited-With Fuel Spray Injection", \href{https://www.sciencedirect.com/science/article/abs/pii/0010218089901314}{Combustion and Flame} 77, pp. 229-240, (1989).


\bibitem{ga10} \label{ga10}
Tambour, Y., A Lagrangian Sectional Approach for Simulating Droplet Size Distribution of Vaporizing Fuel Sprays in a Turbulent Jet, \href{https://www.sciencedirect.com/science/article/abs/pii/0010218085901154}{Combustion and Flame} 61 Issue 1, pp. 15-28,
(1985).


\bibitem{ga11} \label{ga11}
Williams, F.A., Phys. Fluids 1, pp. 541-545, (1958). See also "\href{https://www.routledge.com/Combustion-Theory/Williams/p/book/9780201407778}{Combustion Theory}", 2nd
Edition, The Benjamin/Cummings Publishing: Menlo Park, CA, (1985)


\bibitem{c1} \label{c1}
Boyd, Stephen P.; Vandenberghe, Lieven (2004). \href{https://web.stanford.edu/~boyd/cvxbook/bv_cvxbook.pdf#page=143}{Convex Optimization} page 143 (pdf). Cambridge University Press. p. 129. ISBN 978-0-521-83378-3. 


\bibitem{continuity}  \label{continuity} 
V. Jeyakumar; Alexander M. Rubinov (9 March 2006). Continuous Optimization: Current Trends and Modern Applications. Springer Science $\&$ Business Media. ISBN 978-0-387-26771-5.
\href{https://books.google.co.il/books?id=QePsbLIwwEoC&printsec=frontcover&dq=%22Continuous+optimization%22&hl=en&sa=X&redir_esc=y#v=onepage&q=%22Continuous%20optimization%22&f=false}{Continuous Optimization}


\bibitem{MILP_1} \label{MILP_1}
Javier Larrosa, Albert Oliveras, Enric Rodrıguez-Carbonell (2019), Combinatorial Problem Solving (CPS). 
\href{https://www.cs.upc.edu/~erodri/webpage/cps/theory/lp/milp/slides.pdf}{Mixed Integer Linear Programming}


\bibitem{MILP_2} \label{MILP_2}
Sakawa M. (2002) Genetic Algorithms for Integer Programming. In: Genetic Algorithms and Fuzzy Multiobjective Optimization. \href{https://link.springer.com/chapter/10.1007/978-1-4615-1519-7_5}{Operations Research} / Computer Science Interfaces Series, vol 14. Springer, Boston, MA. 


\bibitem{matlab_problem} \label{matlab_problem}
Matlab Help Center, Solving Mixed Integer GA Optimization Problems [\href{https://www.mathworks.com/help/gads/mixed-integer-optimization.html#bs1clc2}{link}]. Based on Deb, K. (2000). An efficient constraint handling method for genetic algorithms. Computer methods in applied mechanics and engineering, 186(2-4), 311-338 [\href{https://www.sciencedirect.com/science/article/pii/S0045782599003898}{link}].


\bibitem{noise} \label{noise}
Darrel Whitley (1997), A Genetic Algorithm Tutorial, Computer Science Department, Colorado State University, \href{https://web.archive.org/web/20130615042000/http://samizdat.mines.edu/ga_tutorial/ga_tutorial.ps}{Fort Collins}.


\bibitem{HT} \label{HT}
Patankar, S. (2018). Numerical heat transfer and fluid flow. \href{https://catatanstudi.files.wordpress.com/2010/02/numerical-heat-transfer-and-fluid-flow.pdf}{Taylor \& Francis}

\bibitem{linan} \label{linan}
Linan, A. (1974). The asymptotic structure of counterflow diffusion flames for large activation energies. \href{https://www.sciencedirect.com/science/article/abs/pii/0094576574900666}{Acta Astronautica}, 1(7-8), 1007-1039.


\end{thebibliography}
\end{document}